\documentclass{article}

\usepackage[preprint]{tmlr}

\usepackage[utf8]{inputenc}
\usepackage{booktabs}
\usepackage{graphicx}
\usepackage{amsmath}
\usepackage{siunitx}
\usepackage{array}
\usepackage{xcolor}
\usepackage[hidelinks]{hyperref}
\usepackage[expansion=false]{microtype}
\usepackage{tikz}
\usepackage{pgfplots}
\usepackage{float}
\pgfplotsset{compat=1.17}
\usetikzlibrary{arrows.meta, positioning, fit, backgrounds, calc}

\newcommand{\REVIEW}[1]{#1}
\newcolumntype{L}[1]{>{\raggedright\arraybackslash}p{#1}}
\title{Benchmarking Machine-Learning Classifiers for Neoantigen
Immunogenicity Prediction and the Limits of SHAP Interpretability}

\makeatletter
\def\email{\small\it}
\makeatother
\author{\name Ram\'on Gonz\'alez \\
        \addr Mentomy AI, Almer\'ia, Spain \\
        \email ramon@mentomy.com}

\begin{document}
\maketitle

\begin{abstract}
Personalized cancer vaccines require prioritizing, among hundreds of candidate
mutant peptides, the few that will induce a T-cell response. This study asks
what general-purpose machine learning achieves on that task, and whether the
features such models identify as important can be trusted.

We make three contributions. The first is a benchmark of sixty models: twelve
supervised classifier families, each in five configurations, evaluated on the
Immunogenic Tumor Specific Neoantigen database (ITSNdb; 199 experimentally
validated neoantigens, 129 immunogenic and 70 non-immunogenic). All sixty are
compared under one stratified $k$-fold cross-validation protocol, in which the
data used to choose a configuration are never the data used to score it. The
second is an interpretability study that combines SHAP with feature ablation.
The third is a shuffled-label control that tests whether a SHAP value has been
earned.

Two results follow, and both run against current practice. First, the classical
linear models lead the ranking. The gradient-boosted ensembles and neural
networks that dominate recent work bring no advantage. Second, SHAP can fail on
high-cardinality categorical features. Ablation establishes the presenting HLA allele as the
dominant input and the source gene as dispensable. SHAP reports the opposite,
assigning \SIrange{68.7}{81.7}{\percent} of its total value to the source gene
in every model that gives each gene its own weight. Shuffling the gene labels at
random among the peptides resolves the contradiction: the meaningless column
earns the same SHAP value as the real one. That value measures only that ITSNdb
contains 173 genes for 199 peptides. We therefore recommend that SHAP values for
high-cardinality categorical annotations in cohorts of this size always be
reported against such a shuffled-label control, and we describe one in full.
\end{abstract}


\section{Introduction}
\label{sec:intro}

In 1908, accepting the Nobel Prize that recognized his work on immunity, Paul
Ehrlich articulated the idea of the \emph{magic bullet}: an agent that would
recognize what is foreign to the body and destroy it while leaving healthy
tissue untouched \citep{bosch2008}. More than a century later, personalized
cancer vaccines (PCVs) are among the most literal attempts to realize that idea, and
the difficulty has shifted from building the bullet to choosing its target.

Tumors accumulate somatic mutations, a subset of which generate peptide
sequences absent from the normal proteome (the complete set of proteins the
body's healthy cells produce). Such mutated peptides are termed
\emph{neoantigens}. Because they are tumor-specific and have not been subject to
central immune tolerance, neoantigens are attractive targets for therapeutic
vaccination \citep{hu2018,schumacher2015}.
Early-phase trials in melanoma established that personalized neoantigen vaccines
are feasible, safe, and able to induce neoantigen-specific T-cell receptor (TCR)
responses \citep{blass2021,ott2017,sahin2017}, and the strategy is being extended across tumor types
and combined with checkpoint blockade (therapy that releases the molecular
brakes tumors use to switch off T cells).

\begin{figure}[!htb]
\centering
\includegraphics[width=0.86\textwidth]{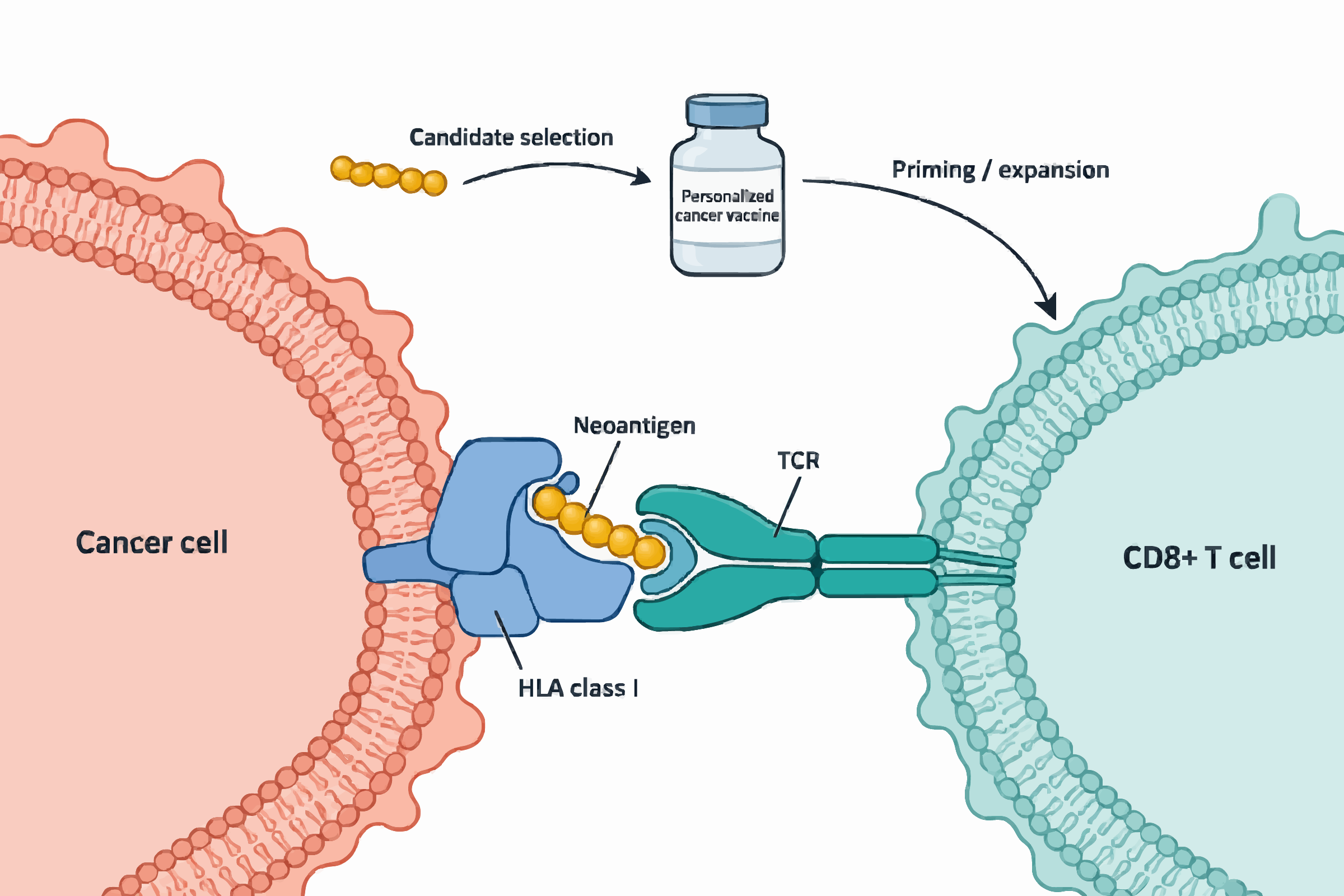}
\caption{Biological context of personalized cancer vaccination and the scope of
this work. A non-silent somatic mutation yields a mutant peptide, which is
processed in the cytosol and loaded onto an HLA class-I molecule for display on
the tumor-cell surface. A CD8$^{+}$ T cell --- a killer cell of the immune
system --- whose T-cell receptor (TCR) engages that peptide--HLA complex with
sufficient affinity can be activated and kill the cell. A personalized vaccine administers a selected subset of such peptides;
these are taken up by dendritic cells, which migrate to the draining lymph nodes
and present them to naive CD8$^{+}$ T cells, priming and clonally expanding the
T-cell populations that recognize them. This paper addresses the
\emph{candidate selection} step, which determines which peptides enter that
subset.}
\label{fig:pcv}
\end{figure}

Figure~\ref{fig:pcv} shows the biological process a personalized cancer vaccine
depends on, and marks the step this paper addresses. A somatic mutation yields a mutant
peptide that, once processed, is displayed on the surface of the tumor cell by
an HLA class-I molecule. A CD8$^{+}$ T cell --- one of the immune system's
killer cells --- whose receptor engages that peptide--HLA complex recognizes the
cell as foreign and can kill it. A
personalized cancer vaccine exploits this by administering a small set of
selected neoantigens to prime and expand the corresponding T-cell populations.
The step labelled \emph{candidate selection} is the computational bottleneck and
the subject of this study: a tumor exome yields hundreds to thousands of mutant
peptides, only a small minority of which are processed, presented and recognized,
while a vaccine formulation has room for only a few dozen: for example, the
individualized mRNA vaccine mRNA-4157/V940, marketed by Moderna as intismeran
autogene and evaluated with pembrolizumab in high-risk resected melanoma,
encodes up to 34 neoantigens per patient \citep{khattak2026}. Existing pipelines rely
heavily on peptide--MHC binding predictors such as the NetMHCpan family \citep{reynisson2020}, but binding is necessary rather than sufficient for
immunogenicity, and the search for features that discriminate immunogenic from
non-immunogenic \emph{binders} remains open.

A systematic benchmark on the curated ITSNdb resource evaluated seven widely
used binding and immunogenicity predictors and found areas under the ROC curve
of only \numrange{0.52}{0.60}, concluding that none could be considered the best
immunogenicity predictor \citep{nibeyro2023}. That benchmark motivates this work
in two ways. First, if purpose-built predictors perform
close to chance, it is worth establishing what general-purpose supervised
classifiers achieve on the same data, under a validation protocol strict enough
that the estimate can be trusted. Second, it makes the question of \emph{which
features carry signal} more pressing than the question of which predictor to
adopt. As \citet{auslander2021} remark, the utility of machine
learning in bioinformatics ultimately depends on the rigor of the entire
pipeline, from careful dataset curation and sequence representation through to
robust evaluation and result interpretability --- a concern this study takes as
its starting point rather than its conclusion.

\paragraph{Contributions.} This paper makes three methodological
contributions, which together yield two results that run against current
practice in this field.

The first is \textbf{a controlled benchmark of sixty machine-learning models}
(Sections~\ref{sec:perf}--\ref{sec:config}): twelve classifier families, each in
five configurations, compared on ITSNdb under a single protocol in which the
data used to choose a configuration are never the data used to score it. The
second is \textbf{an interpretability study combining SHAP with feature
ablation} (Sections~\ref{sec:shapresults} and~\ref{sec:ablationresults}): SHAP
is applied to all twelve families rather than to one, and the full
cross-validation protocol is repeated over four feature sets, so that what each
model \emph{attributes} to a feature can be compared with what that feature
\emph{contributes}. The third is \textbf{a shuffled-label control for SHAP on
categorical features} (Section~\ref{sec:null}): each categorical column is
randomly permuted, preserving its number of categories while destroying any link
to the outcome, and the SHAP analysis is repeated, so that the share a feature
earns under this control is the share it earns from nothing.

These lead to two novel results. \textbf{First, the classical linear models
outperform the more complex ones.} LDA, LR and SGD lead the ranking, while the
gradient-boosted ensembles and neural networks that recent work favours bring no
advantage, and the inner-loop capacity curves explain why: 199 peptides support
very little model complexity. \textbf{Second, SHAP can fail on
categorical features.} The ablation establishes the presenting HLA allele as the
dominant input, removing it costs every one of the twelve families, while the
source gene is dispensable. SHAP reports the opposite, and divides the twelve
families into two opposed groups: every model giving each gene its own weight
ranks the source gene first, with \SIrange{68.7}{81.7}{\percent} of the total,
while every tree-based model assigns it almost nothing (at most
\SI{3.7}{\percent}). Shuffling the gene labels at random reproduces that share
in full, so it measures only that ITSNdb contains 173 genes for 199 peptides. To
our knowledge this is the first report of the effect in neoantigen
immunogenicity prediction, where SHAP over categorical annotations is becoming
standard practice; the control is simple, general, and we describe it in full so
that it can be reused.

\section{Related Work}
\label{sec:related}

This section reviews the state of the art relevant to this study and states, in
each case, how the present work differs. It is organized in three subsections:
the clinical development of personalized cancer vaccines, which defines the
setting; neoantigen discovery and binding prediction, which supplies the
candidates; and the benchmarking of machine-learning models together with the
interpretation of what they learn, which is where this paper contributes.

\subsection{State of the art in PCV clinical trials}
\label{sec:rw-clinical}

The strategy
has moved from proof of concept to multi-arm trials (trials that test
several treatments, or several combinations, against a common control group).
\citet{markosyan2026} survey where the field now stands and identify candidate
selection as a persistent bottleneck rather than a solved problem (the
``selection of neoantigens is key to the success of personalized cancer
vaccines''); \citet{viborg2024} review the DNA-vaccine route to the same goal; and
\citet{mork2022} report a first-in-human trial of the peptide
vaccine EVX-01 with a novel adjuvant in metastatic melanoma. The most advanced
result to date is the randomized phase~IIb KEYNOTE-942 trial, in which the
individualized mRNA vaccine intismeran autogene combined with pembrolizumab
improved recurrence-free survival over pembrolizumab alone in high-risk resected
melanoma \citep{khattak2026}. These studies
define the setting our work serves but do not themselves show how candidates
are chosen computationally.

\subsection{State of the art in neoantigen discovery and binding prediction}
\label{sec:rw-discovery}

\citet{lyu2026} review the
molecular origin, discovery and validation of neoantigens end to end, and
highlight that significant challenges remain, including technical limitations in
reliably identifying all immunogenic neoantigens. \citet{yao2026}
train a random forest on the accept/reject decisions of expert tumor boards
across 33 patients and integrate it into an established toolchain (the pVACtools
suite). That work learns to reproduce expert judgement; ours asks the prior question of
which annotations carry information about immunogenicity at all, and whether the
evidence for them survives scrutiny.

The most mature tools
predict whether a peptide is presented by a given MHC allele. The NetMHCpan
family \citep{reynisson2020} is the current standard, generalizing to unseen
alleles through a pseudo-sequence encoding of the binding pocket. Because
presentation is necessary but not sufficient for a T-cell response, a second
line targets the residual problem directly: DeepHLApan \citep{wu2019},
DeepImmuno \citep{li2021}
and PRIME \citep{schmidt2021}
combine predicted presentation with sequence and amino-acid properties, and more
recent work moves to multimodal architectures that add 3D molecular structures and
TCR--pMHC data \citep{krishnaswamy2026}. All of these learn from the peptide sequence. The
features available in ITSNdb are database annotations instead --- allele, gene,
tumor type, length, mutation position. Our study therefore addresses a different
question. We ask how accurately immunogenicity can be predicted from these
annotations alone, without the sequence, and which of the annotations genuinely
contribute to that accuracy.

\subsection{Benchmarking machine learning for neoantigen prediction and model
interpretability}
\label{sec:rw-bench}

Table~\ref{tab:priorwork} summarizes the data, methods and reported performance
of the machine-learning approaches to this problem, and shows how unevenly the
field reports its results. Experimental benchmarks document how hard the task
remains.

\begin{table}[!htb]
\centering
\caption{Machine-learning approaches to neoantigen immunogenicity prediction:
data, methods and reported performance. \textbf{The values in the last column
are not comparable across rows}, because each was obtained on a different
dataset with a different protocol. The contrast between rows~3 and~6 illustrates the point: tools
reporting ROC-AUC near \num{0.85} on IEDB-derived data fall to
\numrange{0.52}{0.60} when re-evaluated on experimentally validated neoantigens
with both labels. Values are as reported by the authors of each study.}
\label{tab:priorwork}
\footnotesize
\setlength{\tabcolsep}{3.5pt}
\renewcommand{\arraystretch}{1.15}
\begin{tabular}{@{}L{2.35cm}cL{3.2cm}L{3.5cm}L{4.1cm}@{}}
\toprule
Study (tool) & Year & Dataset & ML algorithms & Reported performance \\
\midrule
\citet{wu2019} (DeepHLApan)
 & 2019
 & IEDB: \num{32785} HLA--peptide pairs, \num{7212} immunogenic
 & Recurrent neural network (RNN)
 & No AUC reported for the immunogenicity model; as a filter on NetMHCpan,
   precision \SI{9.6}{\percent} $\rightarrow$ \SI{13.8}{\percent} \\[2pt]

\citet{wells2020} (TESLA)
 & 2020
 & \num{608} screened epitopes; \num{310} in an independent cohort
 & Feature-integrated immunogenicity model
 & Removed \SI{98}{\percent} of non-immunogenic peptides at precision above
   \num{0.70}, at a recall of \num{0.45} \\[2pt]

\citet{li2021} (DeepImmuno)
 & 2021
 & IEDB: \num{8971} peptide--HLA pairs (\num{4059} immunogenic,
   \num{4912} not)
 & DeepImmuno-CNN
 & 10-fold CV: ROC-AUC \num{0.85}, PR-AUC \num{0.81} \\[2pt]

\citet{schmidt2021} (PRIME)
 & 2021
 & Curated dataset (\num{4958} peptides, \num{1282} immunogenic,
   \num{3676} not)
 & Predictor combining antigen presentation with TCR recognition
 & PRIME outperforms known methods on ROC-AUC and PR-AUC \\[2pt]

\citet{muller2023}
 & 2023
 & NCI (112 patients), TESLA (8) and HiTIDE (11), harmonized: \num{46017}
   variants, \num{1781445} neo-peptides, \num{178} immunogenic
 & LR, SVM, XGBoost, CatBoost; LR\,+\,XGBoost voting ensemble
 & LR best; \SI{49.1}{\percent} of immunogenic neo-peptides in the top 20
   (NCI, leave-one-out CV); \SI{30.0}{\percent} more in the top 20 than the
   previous ranking \\[2pt]

\citet{nibeyro2023} (ITSNdb)
 & 2023
 & ITSNdb: \num{199} neoantigens (\num{129} immunogenic, \num{70} not)
 & Evaluation of seven published predictors; no new model
 & ROC-AUC \numrange{0.52}{0.60} across all seven \\[2pt]

\citet{shao2025b} (NeoTImmuML)
 & 2025
 & TumorAgDB2.0: \num{10312} peptides (\num{5156}/\num{5156}); independent
   set of \num{1086}
 & LightGBM, XGBoost, RF, NB, LR, SVC, KNN, MLP; weighted ensemble
 & Best single model (LightGBM) ROC-AUC \num{0.867}, accuracy \num{0.779};
   ensemble \num{0.871}; independent set \num{0.887} \\
\midrule
\textbf{This work}
 & 2026
 & ITSNdb: \num{199} neoantigens (\num{129} immunogenic, \num{70} not)
 & Twelve families in sixty configurations (LR, LDA, SGD, KNN, SVM, MLP, DT,
   RF, ETC, ADA, XGB, LGBM)
 & ROC-AUC \numrange{0.702}{0.817}; balanced accuracy
   \numrange{0.633}{0.760}; nested cross-validation \\
\bottomrule
\end{tabular}
\end{table} The Tumor Neoantigen Selection
Alliance found only 37 of 608 tested peptides (\SI{6}{\percent}) to be
immunogenic \citep{wells2020}. On curated data, \citet{nibeyro2023} evaluated seven published predictors on ITSNdb and obtained
ROC-AUCs of \numrange{0.52}{0.60}, concluding that none could be considered
best. Comparisons of general-purpose classifiers also exist: NeoTImmuML \citep{shao2025b} compares eight algorithms and ensembles the three strongest,
reporting ROC-AUCs near \numrange{0.87}{0.89}. Three differences separate that
work from ours. Its 10{,}312 peptides supplement validated records with
simulated positives filtered on predicted affinity and simulated negatives from
dbSNP variants, so the classes are partly separated by binding affinity by
construction, whereas ITSNdb admits only experimentally validated binding
\emph{and} a validated immunogenicity readout. Its features are physicochemical
descriptors rather than annotations. And its hyperparameters, for both the base
models and the ensemble, were optimized by grid search, whereas we compare five
fixed configurations per family, which makes the role of each parameter legible
rather than leaving it inside an optimizer.

Closest to ours is \citet{muller2023}, who compare logistic
regression, support vector machines, XGBoost and CatBoost on three harmonized
cohorts under stratified five-fold cross-validation and who, as we do, assess
feature importance with SHAP on both a linear and a tree-based model. Three
differences matter. Their categorical features are target-encoded, each category
replaced by its immunogenic rate in the training data, so a column contributes a
single number however many categories it has. Ours are one-hot expanded into one
binary column per category. The artifact we report is caused by that expansion,
which means it cannot occur in their pipeline --- and equally that their design
is untested against it. Their configuration selection is isolated by cohort, a
stronger separation than our within-cohort nested resampling, which we return to
as a limitation in Section~\ref{sec:conclusion}. And where they find SHAP
ranking features differently for logistic regression and for XGBoost and read
that as complementary perspectives, we find the same split across twelve
families and show with a shuffled-label control that on our data it is an
artifact of cardinality --- a divergence that cannot be read either way without
such a control, and we are aware of no study in this literature that reports
one. Their work nonetheless corroborates ours on the first question: at four
orders of magnitude more candidates, they too found logistic regression
strongest.

Interpretability methods are now routine
alongside predictive benchmarking. \citet{ponce2024}
provide a practical guide to SHAP analysis in drug development, working through
linear and logistic regression and then three non-linear frameworks --- random
forest, XGBoost and neural networks --- on a regression and a classification
problem. We extend that approach to twelve families under one protocol, and this
is where our principal contribution lies. Applying SHAP to a single model, as is
usual, would have produced a confident and entirely wrong conclusion about which
feature matters. Comparing twelve families reveals that the answer depends on
the model class, and the shuffled-label control of Section~\ref{sec:null} shows
which of the two answers reflects genuine information in the data and which is
produced by the encoding alone. SHAP is already used in this field, on
these model classes and with this library \citep{muller2023}; what is missing,
and what we supply, is a test of whether a SHAP value for a high-cardinality
categorical annotation has been earned.

\section{Machine-Learning Framework for Immunogenicity Prediction}
\label{sec:framework}

Machine learning is the branch of computer science concerned with algorithms
that learn from data and make predictions --- that is, generalize --- on
observations they have not seen \citep{marsland2015}. Its methods divide broadly
into two types. \emph{Supervised} algorithms are given inputs together with the
correct responses, or targets, and adjust their parameters so as to minimize an
error criterion defined by the difference between those targets and the model
outputs. \emph{Unsupervised} algorithms receive no targets and instead exploit
similarities between inputs in order to group similar inputs together. This
paper focuses on supervised learning, because ITSNdb supplies an
experimentally determined immunogenicity label --- positive or negative --- for
every neoantigen it records, and the task is precisely to reproduce that label
from the available annotations.

\begin{figure}[!htb]
\centering
\includegraphics[width=\textwidth]{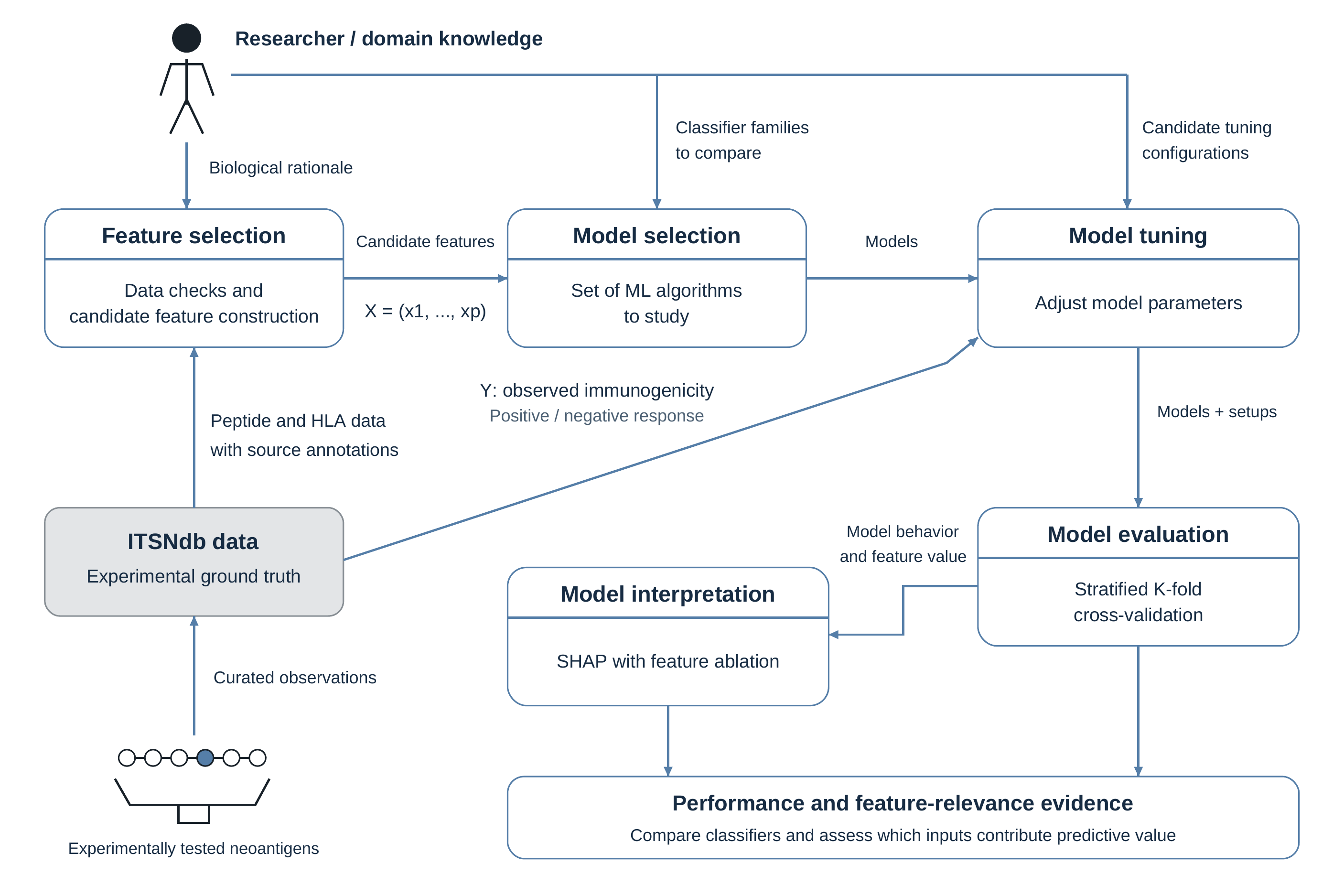}
\caption{Implemented machine-learning framework for neoantigen immunogenicity
prediction. ITSNdb supplies the ground truth; the researcher contributes three
upstream design decisions; and the framework returns the most appropriate models
and configurations together with feature-relevance evidence. The framework is
adapted from the comparison protocol of \citet{gonzalez2019}.}
\label{fig:pipeline}
\end{figure}

Figure~\ref{fig:pipeline} shows the framework implemented to identify the most
appropriate algorithm, and its configuration, among a set of well-known machine
learning algorithms belonging to various categories. The researcher appears in
the schematic as an external actor rather than as a processing step, supplying
three design decisions upstream of any model fitting; representing them this way
localizes the subjective content of the study to three documented points and
makes visible that no decision is taken downstream of the held-out evaluation.

First, the predictor matrix $X=(x_1,\dots,x_p)$ is constructed in the
\textbf{feature selection} stage. Of the columns available in ITSNdb we retained
seven. They were chosen for their biological plausibility rather than by a
statistical screen of the data, so that each one enters the study as a
hypothesis that the later ablation is designed to test
(Section~\ref{sec:features}). The response $Y$ --- the experimentally observed
immunogenicity, positive or negative --- passes directly from the database to the
model-fitting stages without transformation.

Second, in the \textbf{model selection} stage we fixed the set of algorithms to
be compared. Twelve families were chosen, spanning linear models, decision trees,
instance-based learning, ensemble methods, neural networks and kernel machines,
on the basis both of the algorithms prevailing in the current state of the art in
immunogenicity and presentation prediction and of our own previous experience
with an analogous comparison in a different domain \citep{gonzalez2019}
(Section~\ref{sec:families}).

Third, in the \textbf{model tuning} stage each family is instantiated in five
configurations that sweep its principal capacity parameter across the usable
range, from the setting that constrains the model most to the setting that
constrains it least, giving 60 configurations in total.
The swept ranges were fixed by preliminary experimental testing
(Section~\ref{sec:tuning}).

Fourth, having fixed the features, the models and their configurations, the
\textbf{model evaluation} stage estimates the performance of those
configurations by stratified $k$-fold cross-validation \citep{kohavi1995,pedregosa2011}. In $k$-fold cross-validation
the dataset is partitioned into $k$ disjoint subsets of approximately equal size;
each subset serves in turn as the test partition while the remaining $k-1$ are
used for training, so every observation is predicted exactly once by a model that
never saw it during fitting, and the $k$ resulting scores are summarized as a
mean and a standard deviation. \emph{Stratified} $k$-fold additionally requires
every partition to preserve the class prevalence of the complete dataset. That
constraint matters here: with only 70 non-immunogenic records in total, each test
partition contains approximately 14 of them, and unstratified splitting could
easily produce folds whose class composition differs materially, inflating
variance and making the specificity estimates incomparable across folds.

We use $k=5$ for evaluation, and configuration selection is carried out by a
further stratified $k$-fold partition ($k=4$) of each training partition, on
balanced accuracy, so that the data used to choose a family's configuration are
never the data used to score it \citep{varma2006}. Preprocessing is fitted only
on training data within each fold. We refer throughout to the five evaluation
partitions as the \emph{outer folds} and to the selection partition within each
of them as the \emph{inner loop}.

Performance is then reported with the standard metrics for binary classification
under class imbalance --- balanced accuracy, sensitivity and specificity, the F1
score, and the areas under the ROC and precision--recall curves \citep{saito2015,sokolova2009} --- defined in Section~\ref{sec:metrics}.

Finally, a \textbf{model interpretation} stage examines how the fitted models use
their inputs, combining SHAP attribution \citep{lundberg2017} with randomization
controls and with retraining under feature ablation. The pairing is essential to
the results of Section~\ref{sec:ablation}, where the two instruments disagree and
the randomization control determines which of them is informative. The framework
returns the two outputs shown at the foot of Figure~\ref{fig:pipeline}: the most
appropriate models and configurations for this problem, and evidence on which of
the candidate features genuinely contribute predictive value.
\section{Machine-Learning Models and Configurations}
\label{sec:models}

\subsection{ML models}
\label{sec:families}

To characterize what supervised learning can extract from validated neoantigen
annotations, this study compares twelve well-established algorithms drawn from
the principal categories of supervised learning \citep{gonzalez2019}. The intent is deliberately broad
rather than deep. Each family makes different assumptions about the shape of the
relationship between features and outcome --- a hyperplane, a set of nested
rules, a distance, a smooth non-linear surface --- and running all of them lets
us tell apart the conclusions that hold whatever the assumption from those that
merely follow from one family's assumption. That distinction proves decisive for
the interpretability results in Section~\ref{sec:ablation}.

The first category represents \textbf{linear models}, and the algorithms are
Logistic Regression (LR) \citep{hastie2009}, Linear Discriminant Analysis (LDA)
\citep{hastie2009} and a linear classifier fitted by Stochastic
Gradient Descent (SGD) \citep{zhang2004}. Linear means that the logarithm of the class
probability ratio --- the probability that a peptide is immunogenic divided by
the probability that it is not --- is modelled as a weighted sum of the inputs.
Linear methods of this kind are fast and scalable, and they are usually robust
to outliers and to mislabelled observations, which makes them a natural
reference point against which more flexible models have to justify themselves.

The second category represents \textbf{decision trees}, and the algorithm is the
Decision Tree (DT) \citep{mitchell1997}. A decision tree recursively partitions
the feature space using thresholds on single attributes, so the model takes the
form of nested decisions and can represent interactions that a linear model
cannot.
Trees handle categorical and numeric features natively and yield readable rules,
at the cost of high variance: small changes in the training sample can produce
substantially different partitions.

The third category represents \textbf{instance-based learning}, and the
algorithm is $k$-Nearest Neighbors (KNN) \citep{hastie2009}.
Instance-based means that no model is fitted at training time; the training
examples themselves are retained, and a new observation is classified by the
labels of the examples closest to it under a distance metric. The approach assumes nothing
about the shape of the decision boundary, but it depends entirely on the
distance metric being meaningful --- a point we return to when interpreting its
performance.

The fourth category represents \textbf{ensemble methods}, and the algorithms are
Random Forests (RFC) \citep{breiman2001}, Extremely Randomized Trees (ETC)
\citep{geurts2006}, AdaBoost (ADA) \citep{freund1997}, XGBoost (XGB)
\citep{chen2016} and LightGBM (LGBM) \citep{ke2017}. An ensemble combines many
base learners, each individually weak, into a single stronger predictor, and the
two sub-families differ in how the combination is built. In \emph{bagging} the base learners are
trained independently and averaged, which reduces variance; RFC and ETC belong
to this sub-family. In \emph{boosting} they are fitted in sequence, each
correcting the errors of its predecessors, which reduces bias; ADA, XGB and LGBM
belong to this one.

The fifth category represents \textbf{artificial neural networks}, and the
algorithm is the Multi-Layer Perceptron (MLP) \citep{marsland2015}. A multi-layer
perceptron is a feedforward network in which each unit applies a non-linear
activation to a weighted sum of the outputs of the previous layer, so that
composing layers allows arbitrary decision boundaries to be approximated. Its performance relative to
the linear models therefore measures how much non-linear structure the data
actually support.

The sixth and final category represents \textbf{kernel machines}, and the
algorithm is the Support Vector Machine (SVM) \citep{cortes1995}. An SVM seeks
the hyperplane that separates the two classes with the widest possible margin,
determined only by the training points closest to the boundary, the support
vectors.
Replacing inner products with a kernel function allows non-linear boundaries, so
comparing the linear and radial-basis variants tests directly whether
non-linearity helps on these features.

\subsection{Configurations and hyperparameter tuning}
\label{sec:tuning}

Every family above exposes several hyperparameters, and a benchmark that tuned
all of them exhaustively would be neither computationally tractable at 60
configurations nor interpretable in its outcome. We therefore adopted a design in
which, for each family, all settings are fixed at values appropriate to the
problem except \emph{one principal capacity parameter}, which is swept across
five values.

The swept parameter is in each case the one most directly controlling the
capacity--regularization trade-off, and Table~\ref{tab:setups} gives the
complete specification. For LR and SVM it is the inverse regularization strength
$C$, which scales the penalty on the coefficient vector, so that small values
force a smooth, heavily constrained boundary and large values allow the model to
fit the training data closely. For LDA it is the covariance shrinkage
coefficient, which interpolates between the empirical covariance matrix and a
diagonal target; the \texttt{auto} setting estimates it analytically rather than
fixing it. For SGD and MLP it is the L2 penalty $\alpha$, which plays the
inverse role to $C$: larger values mean stronger regularization. For DT, RFC,
ETC and XGB it is the maximum tree depth, which bounds the order of interaction
the model can represent --- a depth of one admits no interactions at all, while
an unbounded depth allows a tree to isolate individual observations. For KNN it
is the neighbourhood size $k$, swept jointly with the Minkowski exponent $p$
that selects the Euclidean ($p=2$) or Manhattan ($p=1$) metric, since the two
interact in a sparse one-hot space. For ADA it is the number of boosting rounds,
which for a fixed learning rate determines how much of the residual structure
the ensemble has had the opportunity to absorb. And for LGBM it is the minimum
number of observations per leaf: raising it limits capacity by refusing any
split that would leave fewer than that many peptides on one side, which prevents
the tree from carving out rules that fit only a handful of observations.

\paragraph{How the ranges were established.} The ranges in
Table~\ref{tab:setups} are not a first guess. We ran an initial grid for each
family, looked at how it behaved across the folds, and moved the grid whenever
the interesting part of the parameter range turned out to lie somewhere else.
We also drew on our own experience of applying this kind of comparison in an
earlier study \citep{gonzalez2019}. Three families --- KNN, MLP and AdaBoost ---
carry no class-weight correction, since their implementations expose no such
parameter or did not benefit from one; this is relevant to interpreting their
results.

\begin{table}[!htb]
\centering
\caption{The five configurations (S1--S5) defined for each classifier family.
``Varied'' is the swept capacity parameter; ``Fixed settings'' lists the values
held constant across that family's configurations.}
\label{tab:setups}
\footnotesize
\setlength{\tabcolsep}{2.6pt}
\begin{tabular}{@{}l l c c c c c L{6.6cm}@{}}
\toprule
Family & Varied & S1 & S2 & S3 & S4 & S5 & Fixed settings \\
\midrule
LR   & $C$ & 0.01 & 0.1 & 1.0 & 10 & 100 &
  balanced class weights; liblinear solver; max\_iter 5000; features standardized \\[1pt]
LDA  & shrink. & 0.1 & 0.25 & 0.5 & 0.75 & auto &
  lsqr solver; equal class priors; features standardized \\[1pt]
SGD  & $\alpha$ & $10^{-5}$ & $10^{-4}$ & $10^{-3}$ & $10^{-2}$ & $10^{-1}$ &
  log loss; L2 penalty; balanced class weights; averaged coefficients; max\_iter 5000; standardized \\[1pt]
KNN  & $(k,p)$ & (5,2) & (9,2) & (15,2) & (9,1) & (15,1) &
  distance weighting; numeric columns standardized, one-hot columns left at 0/1 \\[1pt]
DT   & depth & 2 & 3 & 4 & 5 & 8 &
  min\_samples\_leaf 5; balanced class weights \\[1pt]
SVM  & (kern.,$C$) & (lin,0.1) & (lin,1) & (rbf,0.1) & (rbf,1) & (rbf,10) &
  gamma = scale; balanced class weights; probability estimates enabled; standardized \\[1pt]
MLP  & $\alpha$ & 0.01 & 0.1 & 1.0 & 3.0 & 10 &
  one hidden layer of 16 ReLU units; LBFGS solver; max\_iter 5000; standardized \\[1pt]
RFC  & depth & 3 & 5 & 8 & 12 & none &
  400 trees; min\_samples\_leaf 3; max\_features 0.5; balanced\_subsample weights \\[1pt]
ETC  & depth & 3 & 5 & 8 & 12 & none &
  400 trees; min\_samples\_leaf 3; max\_features 0.5; balanced class weights \\[1pt]
ADA  & rounds & 50 & 100 & 200 & 400 & 600 &
  depth-1 decision stumps; learning rate 0.1; SAMME algorithm \\[1pt]
XGB  & depth & 1 & 2 & 3 & 4 & 5 &
  300 rounds; learning rate 0.03; subsample 0.8; colsample\_bytree 0.8; $\lambda_{L2}=5$; scale\_pos\_weight 0.54 \\[1pt]
LGBM & min.\ leaf & 3 & 5 & 10 & 15 & 20 &
  200 rounds; 7 leaves; learning rate 0.03; $\lambda_{L2}=1$; \texttt{is\_unbalance} enabled \\
\bottomrule
\end{tabular}
\end{table}

\section{Dataset and Features}
\label{sec:data}

\subsection{The ITSNdb resource}

We used the Immunogenic Tumor Specific Neoantigen database (ITSNdb)
\citep{nibeyro2023}, assembled to address weaknesses that had limited earlier
immunogenicity benchmarks: the conflation of unvalidated with true negatives, and
the mixing of predicted with experimentally confirmed presentation. It is
manually curated from more than 70 publications, of which 45 contributed
entries that met its strict inclusion criteria. Each entry derives from a non-silent somatic
single-nucleotide variant whose wild-type sequence is confirmed in the referenced
protein; has experimentally validated MHC class-I binding by mass spectrometry or
competition assay; and carries an experimentally validated positive or negative
immunogenicity readout from assays such as tetramer titration or
IFN-$\gamma$/TNF ELISPOT.

The cohort comprises \textbf{199 neoantigens}, all nine- and ten-mers derived
from single-nucleotide variants, each paired with its wild-type counterpart:
\textbf{129 immunogenic} and \textbf{70 non-immunogenic}, a positive-class
prevalence of \num{0.648}. The imbalance favours the \emph{immunogenic} class,
which is why performance is estimated by stratified $k$-fold cross-validation
(Section~\ref{sec:framework}; its consequences are visible in
Section~\ref{sec:perf}): stratification keeps that prevalence constant
across folds, so the 70 non-immunogenic records are distributed evenly rather
than concentrating in a few partitions.

Because every ITSNdb peptide is already a validated binder, the learning task is
not ``does this peptide bind?'' but the harder residual question ``given that it
binds and is presented, is it immunogenic?''. A negative label means no response
was detected under the assay conditions tested; it does not establish universal
non-immunogenicity.

\subsection{Feature selection}
\label{sec:features}

\begin{figure}[!htb]
\centering
\includegraphics[width=\textwidth]{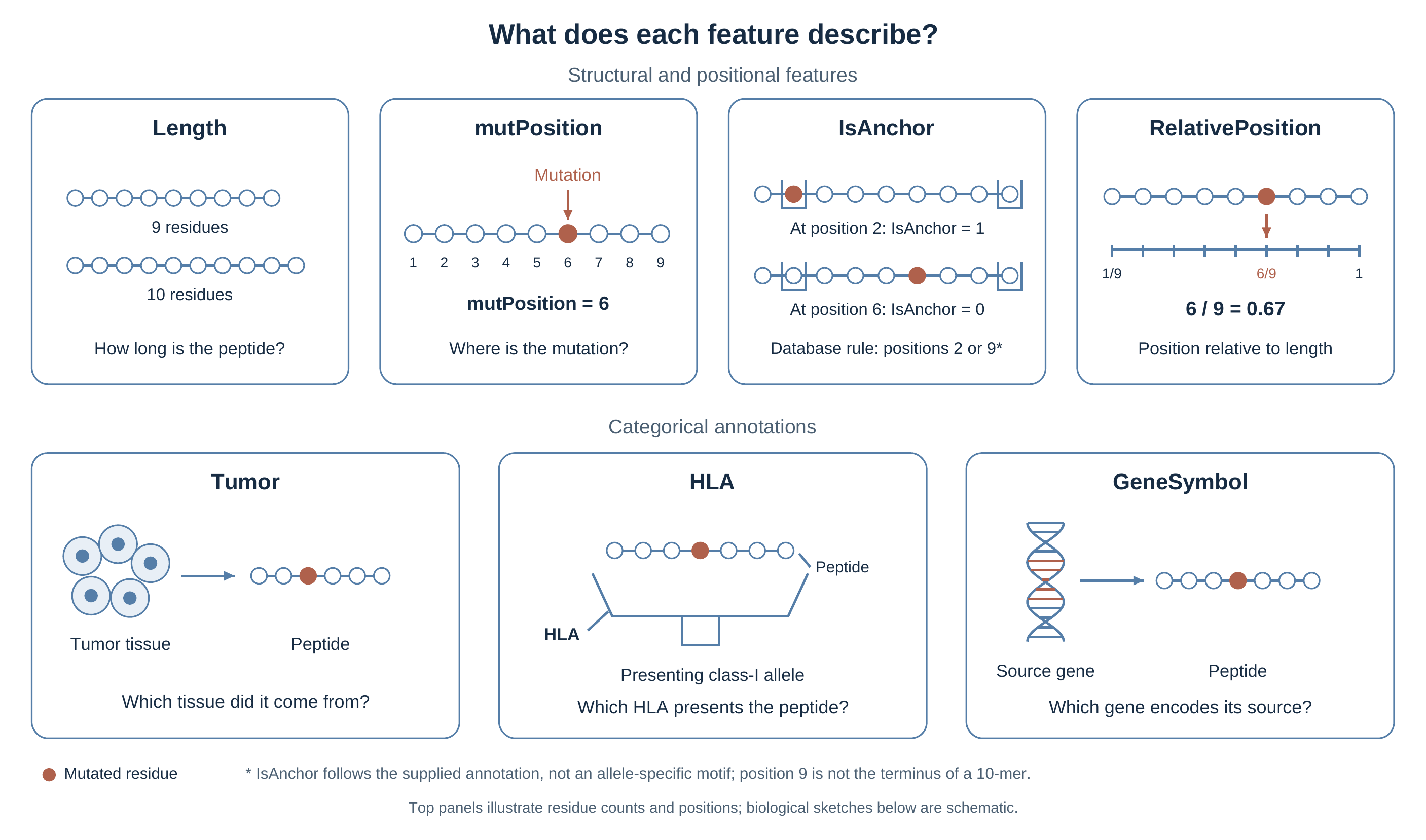}
\caption{Schematic description of the seven selected features. The upper row
covers the structural and positional features, the lower row the categorical
annotations.
The mutated residue is marked in a contrasting colour. The biological sketches
are schematic.}
\label{fig:features}
\end{figure}

Seven features were assembled from the database annotations: two numeric, two
derived and three categorical. Table~\ref{tab:features} summarizes them together
with the distribution of each across the 199 entries, and
Figure~\ref{fig:features} illustrates what each one describes.

\paragraph{Structural and positional features.} \texttt{Length} is the peptide
length in residues, which determines how the peptide is accommodated in the
MHC class-I binding groove and is a standard input to binding and immunogenicity
predictors \citep{reynisson2020,wu2019}.
ITSNdb contains only nine- and ten-mers --- \textbf{149 of length 9} and
\textbf{50 of length 10} --- so this feature is effectively binary here and its
contribution must be read as a single 9-mer-versus-10-mer contrast rather than
as a general length effect. \texttt{mutPosition} is the position of the mutated
residue within the peptide, ranging from 1 to 10 and distributed fairly evenly
across the sequence, with mild peaks at positions 2 (27 entries), 8 (28) and
9 (27). Its relevance is that a mutation at a position facing the T-cell
receptor can create an epitope recognized as foreign while leaving presentation
unchanged, whereas one at a buried position facing the MHC molecule may alter
binding without changing what the receptor sees; Capietto et al.\ showed
experimentally that mutation position is an important determinant of
immunogenicity \citep{capietto2020}. \texttt{IsAnchor} indicates whether the
mutation falls at an \emph{anchor} position --- one of the few positions whose
side chain is buried in the MHC groove and holds the peptide in place, rather
than pointing outwards towards the T-cell receptor. ITSNdb marks a mutation as
anchor when it lies at position 2 or 9, giving \textbf{54 anchor} and
\textbf{145 non-anchor} entries. \texttt{RelativePosition} is \texttt{mutPosition} divided by
\texttt{Length}, which places mutations on a common scale across the two peptide
lengths; it adds no information beyond the other two jointly, but supplies their
interaction in a form additive models can use directly. An additive model such
as logistic regression assigns one weight per feature and sums the results, so
it cannot by itself represent the fact that position 9 means something different
in a nine-mer than in a ten-mer. Supplying the ratio as its own column states
that relationship explicitly, so the model can use it without having to discover
it.

\paragraph{Categorical annotations.} \texttt{HLA} is the human leukocyte
antigen, the molecule that presents intracellular peptides on the cell surface
to CD8$^{+}$ T cells. Presentation is allele-restricted: the allele determines
which peptides can be displayed at all, and all contemporary predictors are
explicitly allele-aware \citep{li2021,reynisson2020,wu2019}.
The cohort covers \textbf{31 alleles}, but usage is heavily concentrated:
HLA-A*02:01 alone accounts for 80 entries (\SI{40.2}{\percent}), followed by
HLA-B*27:05 and HLA-A*03:01 with 21 each, so the remaining 28 alleles share 77
entries. \texttt{Tumor} is the tumor type from which the neoantigen was
identified. Sixteen types are present and the distribution is likewise skewed:
melanoma accounts for 146 entries (\SI{73.4}{\percent}), followed by breast
(15), lung (7) and colorectal (6), with the remaining twelve types --- among
them head and neck, cervical, ovarian, glioblastoma, renal, bladder and
pancreatic --- contributing four entries or fewer each.
\texttt{GeneSymbol} is the gene from which the peptide derives. It is by far the
highest-cardinality feature: \textbf{173 distinct genes across 199 entries},
that is \num{1.15} entries per gene, with only TP53 (6), GAS7 (3) and KRAS (3)
appearing more than twice. Its cardinality turns out to matter for the interpretability
analysis of Section~\ref{sec:ablation}.

\begin{table}[!htb]
\centering
\caption{The seven selected features, their role, and their distribution over the
199 ITSNdb entries. Categorical features were one-hot encoded within each
training fold; the number of distinct values each takes is the cardinality that
Section~\ref{sec:ablation} shows to matter for attribution.}
\label{tab:features}
\small
\setlength{\tabcolsep}{4pt}
\begin{tabular}{@{}l l L{4.2cm} L{5.0cm}@{}}
\toprule
Feature & Type & Role & Distribution in ITSNdb ($n=199$) \\
\midrule
Length           & numeric     & Fit within the MHC-I groove
                 & 2 values: 9 ($n=149$), 10 ($n=50$) \\[1pt]
mutPosition      & numeric     & TCR-facing versus MHC-facing mutation \citep{capietto2020}
                 & 10 values, 1--10; peaks at 8 ($n=28$), 2 and 9 ($n=27$) \\[1pt]
IsAnchor         & binary      & Mutation buried in the MHC groove
                 & anchor ($n=54$), non-anchor ($n=145$) \\[1pt]
RelativePosition & numeric     & Length-normalized mutation location
                 & derived from \texttt{Length} and \texttt{mutPosition} \\[1pt]
Tumor            & categorical & Tumor-type context; cohort structure
                 & 16 types; melanoma ($n=146$), breast (15), lung (7),
                   colorectal (6) \\[1pt]
HLA              & categorical & Presenting allele \citep{reynisson2020}
                 & 31 alleles; A*02:01 ($n=80$), B*27:05 and A*03:01 (21 each) \\[1pt]
GeneSymbol       & categorical & Source gene; indirect and cohort-specific
                 & 173 genes, \num{1.15} entries per gene; max TP53 ($n=6$) \\
\bottomrule
\end{tabular}
\end{table}

ITSNdb also provides the wild-type and mutated peptide sequences themselves. We
tested them as additional categorical features and excluded them: with 199
peptides the sequences are almost all distinct, so a sequence seen in the test
fold never appeared in training and contributes nothing to the prediction. This is the same effect, in its most extreme form, that
Section~\ref{sec:null} documents for \texttt{GeneSymbol}.

\section{Model Evaluation and Model Interpretability}
\label{sec:evalinterp}

\subsection{Model evaluation}
\label{sec:metrics}

Model evaluation is the process of assessing the performance of a machine
learning model using specific metrics, whereas model selection involves
comparing different models in order to choose the one that performs best
according to those metrics \citep{geron2019,hastie2009,kohavi1995}. In this study we use a
handful of the most widely known metrics in the machine learning literature,
chosen because they remain informative on imbalanced datasets such as ours \citep{saito2015,sokolova2009}. Throughout, the immunogenic peptide is the
positive class, and we write TP, TN, FP and FN for the counts of true positives,
true negatives, false positives and false negatives on a test fold.

\textbf{Accuracy} is the proportion of peptides classified correctly,
$(\text{TP}+\text{TN})/(\text{TP}+\text{TN}+\text{FP}+\text{FN})$. It is the most
intuitive metric but the least useful one here: with a positive-class prevalence
of \num{0.648}, a model that labels every peptide immunogenic already achieves an
accuracy of \num{0.648} while identifying no non-immunogenic peptide at all.

\textbf{Sensitivity} (Sn), also called recall or the true positive rate, is the
proportion of genuinely immunogenic peptides the model recovers,
$\text{Sn}=\text{TP}/(\text{TP}+\text{FN})$. \textbf{Specificity} (Sp), the true
negative rate, is the corresponding proportion for the other class,
$\text{Sp}=\text{TN}/(\text{TN}+\text{FP})$. The two are reported as a pair because they
trade against each other: a model can always raise one by sacrificing the other,
and in this application they carry different costs, since a false positive
occupies a slot in a vaccine formulation.

Because accuracy is dominated by the majority class, we take \textbf{balanced
accuracy} as the primary metric and as the inner-loop selection criterion. It is
the unweighted mean of the two rates above, $\text{balanced accuracy} =
\tfrac{1}{2}(\text{Sn}+\text{Sp})$, so each class contributes equally regardless
of how many peptides it contains.
The constant predictor that scores \num{0.648} on accuracy scores exactly
\num{0.5} here, which is what makes the metric readable: \num{0.5} is chance and
\num{1.0} is perfect, whatever the class balance.

The \textbf{F1 score}, reported in Table~\ref{tab:fullmodels}, is the harmonic
mean of precision ($\text{TP}/(\text{TP}+\text{FP})$) and sensitivity. Both of those quantities
count only immunogenic peptides, so true negatives never enter the formula at
all. A model that labels almost every peptide immunogenic therefore scores well
on F1 while identifying none of the non-immunogenic ones --- which is precisely
the failure mode of this cohort, where the immunogenic class is already the
majority. We report F1 for comparability with other studies but do not use it to
rank models.

Two metrics matter especially in the context of personalized cancer vaccines,
ROC-AUC and PR-AUC. The metrics above are computed from hard predictions at a
fixed decision threshold, whereas a vaccine pipeline does not classify peptides
one by one: it ranks all the candidates from a tumor and takes the few that fit
in the formulation. What matters there is the quality of the ordering, not where
a threshold happens to fall, and these two measure exactly that.
\textbf{ROC-AUC} is the area under the curve of sensitivity against
$1-\text{specificity}$. It has a direct reading: take one immunogenic and one
non-immunogenic peptide at random, and ROC-AUC is the probability that the model
gives the immunogenic one the higher score. A model that ranks at random is
right half the time, so the chance baseline is fixed at \num{0.5} whatever the
class balance. \textbf{PR-AUC} is the area under the
precision--recall curve, which is more sensitive than ROC-AUC to performance on
the minority class \citep{saito2015}. Its baseline is \emph{not} \num{0.5} but the
positive-class prevalence, \num{0.648}, so PR-AUC values must be read against
that floor rather than as standalone figures.

\subsection{Model interpretation: SHAP}
\label{sec:shapmethod}

Model interpretability is the study of \emph{why} a model predicts what it
predicts: which inputs drive its output, by how much, and in which direction. It
matters here for two reasons. The first is practical. The output of a
personalized cancer vaccine pipeline is not a score but a decision about which
handful of peptides to synthesize for a particular patient, and that decision is
taken by people who must be able to justify it. The second is specific to small
curated cohorts such as this one: interpretability is the means by which one
discovers that a model is right for the wrong reason --- that it is keyed on an
annotation reflecting how the cohort was assembled rather than on a property of
the peptide. Both reasons motivate examining what the fitted models have
learned, not only how well they score.

SHAP (SHapley Additive exPlanations) is a model interpretability methodology
rooted in game theory, designed to quantify the contribution of each feature to
the model's predictions \citep{bai2026,lundberg2017}. By calculating the SHAP
values for all samples, an overall importance ranking of the features can be
derived. A higher SHAP value associated with a feature signifies a more
substantial impact of that feature on the model's predictions.

For each of the twelve families we proceeded as follows.

\begin{enumerate}
  \item \textbf{Select the representative configuration.} Identify the
  configuration selected most frequently by the inner loop across the five outer
  folds (Figure~\ref{fig:selfreq}), so that the model explained is the one the
  protocol would have chosen.
  \item \textbf{Refit on the complete dataset.} Fit that configuration on all
  199 observations. This model is used only for interpretation, never for
  performance estimation, which comes from the outer folds.
  \item \textbf{Compute SHAP values.} Apply SHAP's \textbf{permutation}
  explainer to the function mapping inputs to the predicted probability of the
  positive class, with a background sample of 50 rows and
  $\texttt{max\_evals} = 10(2M+1)$ for $M = 7$ features. Masking operates on the
  \textbf{seven original columns}, with categorical features represented as
  integer codes and converted back to their labels inside the prediction
  wrapper; the model's own pipeline performs one-hot encoding downstream.
 
  \item \textbf{Compute the mean absolute SHAP value per feature.}
  $\frac{1}{N}\sum_i |\phi_{ij}|$ for feature $j$ over the $N$ observations.
  This is a global measure of feature importance: a larger mean absolute SHAP
  value means a greater average contribution of that feature to the model's
  predictions.
  \item \textbf{Inspect the distribution of the attributions.} The beeswarm
  summary plot shows one row per feature, ranked by mean absolute SHAP value,
  with one point per observation positioned by its SHAP value and coloured by
  the value of that feature \citep{ponce2024}. A single display therefore conveys
  which features matter, how large their effects are, and in which direction a
  high or low feature value pushes the prediction --- information that the mean
  absolute value alone collapses.
  \item \textbf{Check the attribution against a null.} A feature with many
  categories can earn a high mean absolute SHAP value even when it carries no
  information, because the model sees too few peptides per category to learn
  anything reliable from it --- \texttt{GeneSymbol} has 173 genes across 199
  peptides. To detect this we shuffle the categorical columns at random, which
  keeps the number of categories but removes any link to immunogenicity, then
  refit the model and repeat steps 2--4. Five shuffles give the attribution a
  feature earns from noise alone, a \textbf{shuffled-label control}, against
  which we report each feature's real share, its null share and the ratio
  between them; a ratio near 1 means the feature earns no more than noise.
\end{enumerate}

\section{Results}
\label{sec:results}

\subsection{Software implementation}
\label{sec:software}

All software developed for this study was written in Python. The twelve
classifier families, the one-hot encoding of the categorical features, and the
stratified $k$-fold partitioning were implemented
with scikit-learn \citep{geron2019,pedregosa2011}, which also supplied ten of the twelve
algorithms; the remaining two, XGB and LGBM, were taken from the dedicated
\texttt{xgboost} \citep{chen2016} and \texttt{lightgbm} \citep{ke2017} libraries.
Model evaluation used the \texttt{sklearn.metrics} package for every metric
defined in Section~\ref{sec:metrics}, and the interpretability analysis of
Section~\ref{sec:shapmethod} used the \texttt{shap} package \citep{lundberg2017}.
Auxiliary data handling and plotting relied on pandas, NumPy and Matplotlib.

All results reported in this paper were produced with Python 3.10.12,
scikit-learn 1.4.2, XGBoost 3.2.0, LightGBM 4.7.0, SHAP 0.49.1, pandas 2.2.3,
NumPy 1.26.4 and Matplotlib 3.9.0.

\subsection{Model selection: classifier performance}
\label{sec:perf}

This section analyses the experimental results obtained. We compare the
performance of the twelve families of machine learning algorithms, each taken at
the configuration selected for it by the inner loop, on the held-out testing
partitions of the outer folds. In this first experiment all seven features
described in Section~\ref{sec:features} are used.

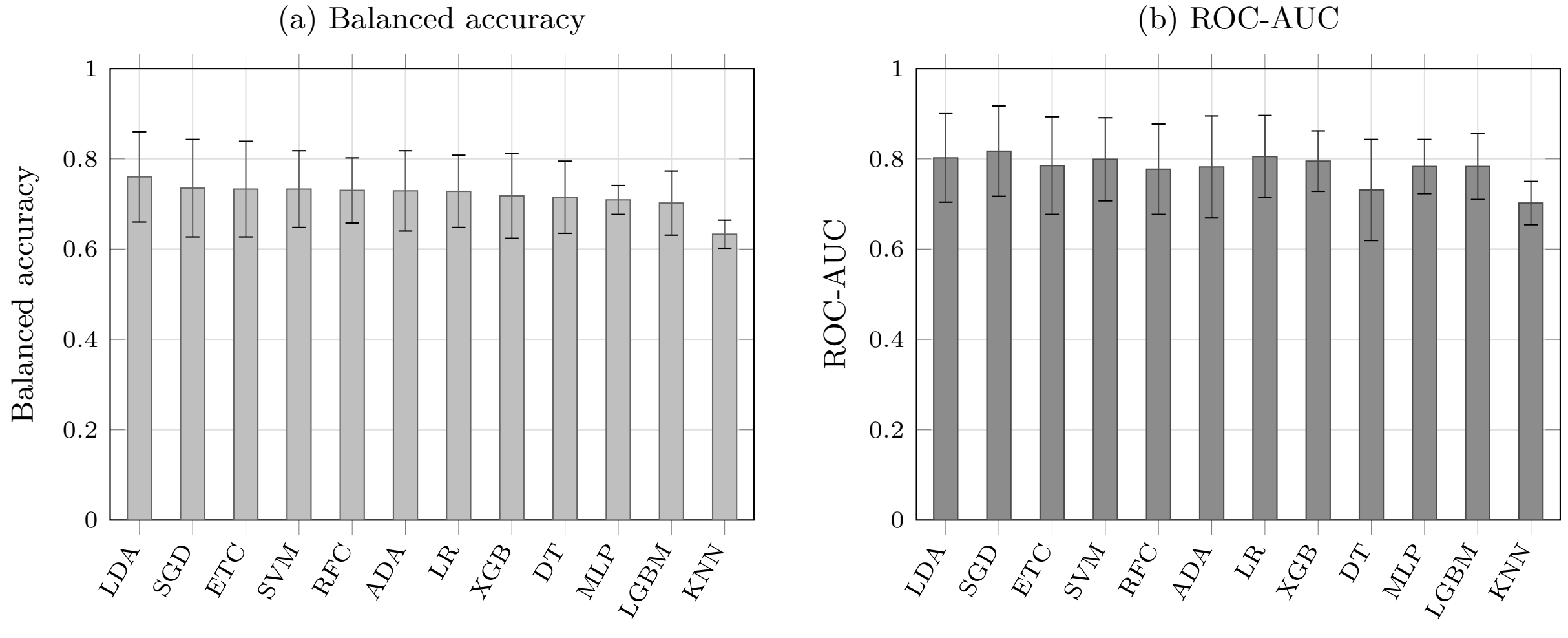
\begin{figure}[!htb]
\centering
\begin{tikzpicture}
\begin{axis}[
  width=0.495\textwidth, height=6.2cm,
  ybar, bar width=7pt,
  symbolic x coords={LDA,SGD,ETC,SVM,RFC,ADA,LR,XGB,DT,MLP,LGBM,KNN},
  xtick=data, x tick label style={rotate=60, anchor=east, font=\scriptsize},
  ymin=0, ymax=1, ytick={0,0.2,0.4,0.6,0.8,1},
  ylabel={Balanced accuracy},
  ylabel style={font=\small}, yticklabel style={font=\scriptsize},
  title={(a) Balanced accuracy}, title style={font=\small},
  grid=major, grid style={black!12},
  error bars/y dir=both, error bars/y explicit,
  enlarge x limits=0.05,
]
\addplot[fill=black!25, draw=black!60] coordinates {
 (LDA,0.760) +- (0,0.100) (SGD,0.735) +- (0,0.108) (ETC,0.733) +- (0,0.106)
 (SVM,0.733) +- (0,0.085) (RFC,0.730) +- (0,0.072) (ADA,0.729) +- (0,0.089)
 (LR,0.728)  +- (0,0.080) (XGB,0.718) +- (0,0.094) (DT,0.715)  +- (0,0.080)
 (MLP,0.709) +- (0,0.032) (LGBM,0.702)+- (0,0.071) (KNN,0.633) +- (0,0.031)};
\end{axis}
\begin{scope}[xshift=0.5\textwidth]
\begin{axis}[
  width=0.495\textwidth, height=6.2cm,
  ybar, bar width=7pt,
  symbolic x coords={LDA,SGD,ETC,SVM,RFC,ADA,LR,XGB,DT,MLP,LGBM,KNN},
  xtick=data, x tick label style={rotate=60, anchor=east, font=\scriptsize},
  ymin=0, ymax=1, ytick={0,0.2,0.4,0.6,0.8,1},
  ylabel={ROC-AUC},
  ylabel style={font=\small}, yticklabel style={font=\scriptsize},
  title={(b) ROC-AUC}, title style={font=\small},
  grid=major, grid style={black!12},
  error bars/y dir=both, error bars/y explicit,
  enlarge x limits=0.05,
]
\addplot[fill=black!45, draw=black!70] coordinates {
 (LDA,0.802) +- (0,0.098) (SGD,0.817) +- (0,0.100) (ETC,0.785) +- (0,0.108)
 (SVM,0.799) +- (0,0.092) (RFC,0.777) +- (0,0.100) (ADA,0.782) +- (0,0.113)
 (LR,0.805)  +- (0,0.091) (XGB,0.795) +- (0,0.067) (DT,0.731)  +- (0,0.112)
 (MLP,0.783) +- (0,0.060) (LGBM,0.783)+- (0,0.073) (KNN,0.702) +- (0,0.048)};
\end{axis}
\end{scope}
\end{tikzpicture}
\caption{Outer-fold performance of the twelve classifier families on the full feature
set: (a) balanced accuracy and (b) ROC-AUC. Bars are means over the five outer
folds; whiskers are $\pm 1$ standard deviation. Families are ordered by balanced
accuracy in both graphs.}
\label{fig:perf}
\end{figure}

Figure~\ref{fig:perf} shows the two metrics of record in this study, balanced
accuracy and ROC-AUC, averaged over the five outer folds, with whiskers giving
the fold-to-fold standard deviation. The first conclusion is a reversal of the
usual expectation. The modern methods that dominate the recent PCV literature
--- the gradient-boosted ensembles XGB and LGBM, the decision tree and the
multi-layer perceptron --- obtain only moderate performance on both metrics,
with balanced accuracies between \num{0.702} and \num{0.718}. The classical
linear methods lead instead: LDA attains the highest balanced accuracy
($0.760 \pm 0.100$) and SGD the highest ROC-AUC ($0.817 \pm 0.100$), with LR
close behind on both. On data of this size and with these features, the simpler
hypothesis class is the better one.

\begin{figure}[!htb]
\centering
\begin{tikzpicture}
\begin{axis}[
  width=0.80\textwidth, height=7.4cm,
  xlabel={Sensitivity (more immunogenic peptides correctly found)},
  ylabel={Specificity (fewer non-immunogenic wrongly selected)},
  xlabel style={font=\small}, ylabel style={font=\small},
  xticklabel style={font=\scriptsize}, yticklabel style={font=\scriptsize},
  xmin=0.74, xmax=0.95, ymin=0.39, ymax=0.70,
  grid=major, grid style={black!12},
]
\foreach \b in {0.65,0.70,0.75} {
  \edef\temp{\noexpand\addplot[dashed, black!45, thin, domain=0.74:0.95,
     samples=2, forget plot] {2*\b - x};}
  \temp
}
\node[font=\scriptsize, text=black!50, rotate=-27] at (axis cs:0.787,0.525) {BA = 0.65};
\node[font=\scriptsize, text=black!50, rotate=-27] at (axis cs:0.926,0.482) {BA = 0.70};
\node[font=\scriptsize, text=black!50, rotate=-27] at (axis cs:0.936,0.572) {BA = 0.75};
\addplot[only marks, mark=*, mark size=1.9pt, draw=black!70, fill=black!55,
         nodes near coords, point meta=explicit symbolic,
         every node near coord/.append style={font=\scriptsize, text=black!85,
           anchor=south west, xshift=1pt, yshift=0pt}]
  coordinates {
   (0.876,0.643) [LDA] (0.899,0.571) [SGD] (0.852,0.614) [ETC]
   (0.838,0.629) [SVM] (0.860,0.600) [RFC] (0.914,0.543) [ADA]
   (0.814,0.643) [LR]  (0.822,0.614) [XGB] (0.845,0.586) [DT]
   (0.861,0.557) [MLP] (0.776,0.629) [LGBM] (0.838,0.429) [KNN]};
\end{axis}
\end{tikzpicture}
\caption{Mean sensitivity against mean specificity over the five outer folds, full
feature set. Dashed diagonals are contours of constant balanced accuracy;
positions closer to the upper-right corner are better on both classes.}
\label{fig:snsp}
\end{figure}
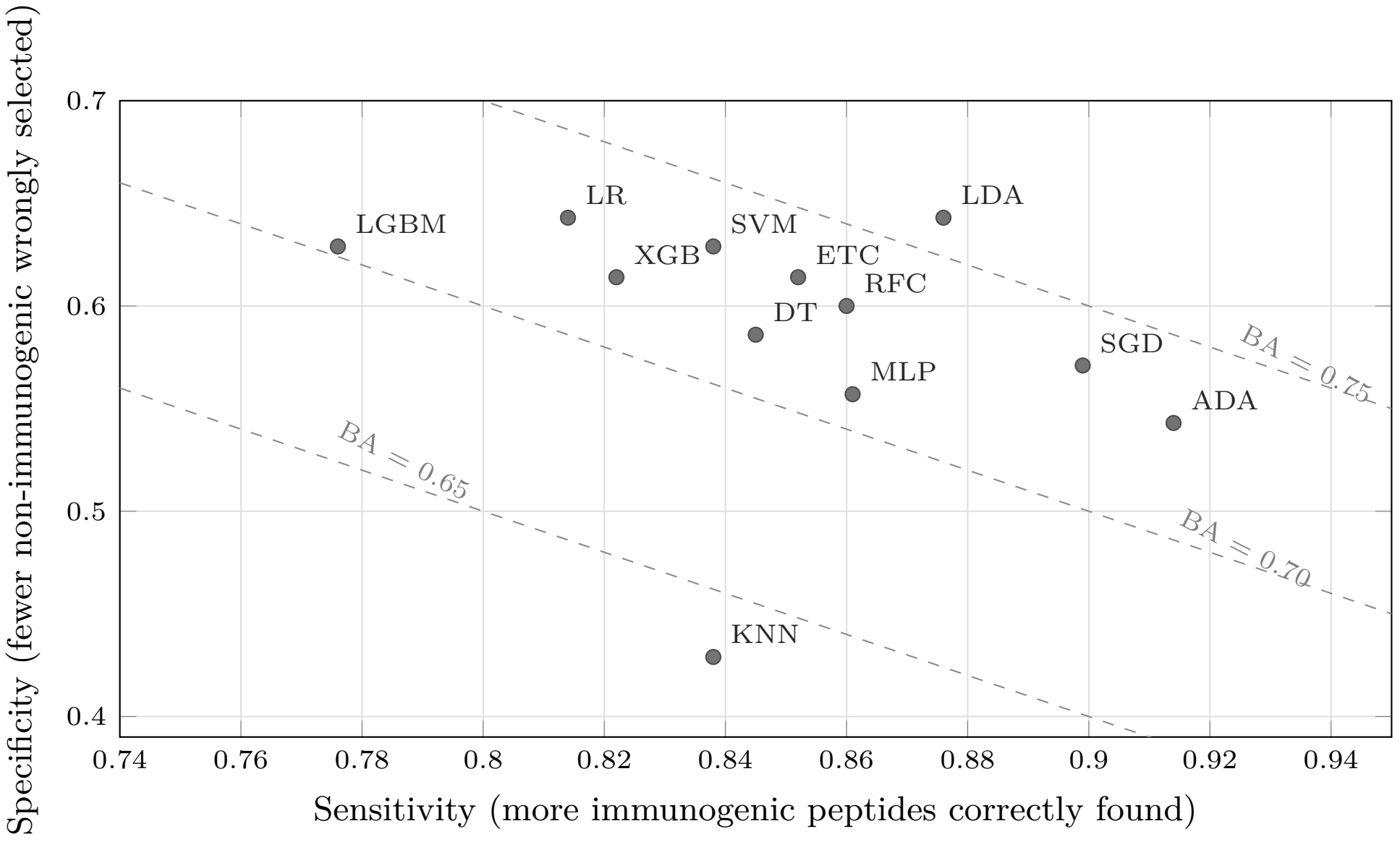

Figure~\ref{fig:snsp} offers a second and more informative comparison, plotting
mean sensitivity against mean specificity; in this graph the closer a family
lies to the upper-right corner, the better it performs on both classes at once.
Two things are visible. LDA sits closest to that corner, confirming its
advantage in Figure~\ref{fig:perf} and showing that the advantage comes from
handling both classes well rather than from favouring one. The tree-based
methods DT, ETC and RFC cluster in the centre of the plot, balanced between the
two error types but dominated by LDA on both axes. ADA and SGD lie towards the
high-sensitivity extreme, buying recall of immunogenic peptides at the cost of
specificity, while KNN is separated from every other family by roughly
\num{0.11} of specificity. The dashed diagonals are contours of equal balanced
accuracy, so families lying on the same line differ in operating point rather
than in overall quality: the choice between them is an application decision
about the relative cost of the two error types, and for vaccine design, where a
false positive occupies a slot in the formulation, the higher-specificity
operating point (fewer non-immunogenic peptides wrongly selected) is
preferable.

Finally, Table~\ref{tab:fullmodels} reports every metric analysed in this study
for all twelve families, again averaged over the five testing partitions. Three
results stand out. First, LDA leads on three of the seven metrics --- balanced
accuracy, accuracy (\num{0.794}) and F1 (\num{0.847}) --- while SGD leads on
both ranking metrics, ROC-AUC (\num{0.817}) and PR-AUC (\num{0.901}); no single
family is best everywhere. Second, and the most consistent feature of the table,
every family shares the same asymmetry: sensitivity spans
\numrange{0.776}{0.914} while specificity spans only \numrange{0.429}{0.643}.
All twelve identify immunogenic peptides considerably more reliably than
non-immunogenic ones. Two properties of the data explain this: the negative
class is scarce, at 70 records or roughly 14 per testing partition, so its
decision region is estimated from few examples; and it is intrinsically
heterogeneous, since ITSNdb negatives are validated binders that failed to
elicit a response.
Third, the PR-AUC column must be read against its prevalence floor of
\num{0.648} rather than against \num{0.5}: on that basis the best PR-AUC
represents a gain of \num{0.253} over a trivial predictor and the worst, KNN's,
only \num{0.148}.

\begin{table}[!htb]
\centering
\caption{Outer-fold performance on the full feature set, mean\,$\pm$\,standard
deviation over the five outer folds, ordered by balanced accuracy. Best value per
column in bold; LDA and LR tie on specificity. Sn: sensitivity; Sp: specificity.}
\label{tab:fullmodels}
\scriptsize
\setlength{\tabcolsep}{4pt}
\begin{tabular}{lccccccc}
\toprule
Family & Accuracy & Balanced acc. & Sn & Sp & F1 & ROC-AUC & PR-AUC \\
\midrule
LDA & \textbf{0.794\,$\pm$\,0.082} & \textbf{0.760\,$\pm$\,0.100} & 0.876\,$\pm$\,0.063 & \textbf{0.643\,$\pm$\,0.175} & \textbf{0.847\,$\pm$\,0.059} & 0.802\,$\pm$\,0.098 & 0.880\,$\pm$\,0.075 \\
SGD & 0.784\,$\pm$\,0.087 & 0.735\,$\pm$\,0.108 & 0.899\,$\pm$\,0.043 & 0.571\,$\pm$\,0.182 & 0.845\,$\pm$\,0.060 & \textbf{0.817\,$\pm$\,0.100} & \textbf{0.901\,$\pm$\,0.067} \\
ETC & 0.768\,$\pm$\,0.083 & 0.733\,$\pm$\,0.106 & 0.852\,$\pm$\,0.052 & 0.614\,$\pm$\,0.193 & 0.828\,$\pm$\,0.059 & 0.785\,$\pm$\,0.108 & 0.856\,$\pm$\,0.079 \\
SVM & 0.764\,$\pm$\,0.067 & 0.733\,$\pm$\,0.085 & 0.838\,$\pm$\,0.073 & 0.629\,$\pm$\,0.171 & 0.821\,$\pm$\,0.051 & 0.799\,$\pm$\,0.092 & 0.880\,$\pm$\,0.067 \\
RFC & 0.769\,$\pm$\,0.061 & 0.730\,$\pm$\,0.072 & 0.860\,$\pm$\,0.070 & 0.600\,$\pm$\,0.139 & 0.828\,$\pm$\,0.045 & 0.777\,$\pm$\,0.100 & 0.848\,$\pm$\,0.081 \\
ADA & 0.784\,$\pm$\,0.077 & 0.729\,$\pm$\,0.089 & \textbf{0.914\,$\pm$\,0.051} & 0.543\,$\pm$\,0.130 & 0.846\,$\pm$\,0.054 & 0.782\,$\pm$\,0.113 & 0.837\,$\pm$\,0.084 \\
LR  & 0.753\,$\pm$\,0.053 & 0.728\,$\pm$\,0.080 & 0.814\,$\pm$\,0.033 & \textbf{0.643\,$\pm$\,0.175} & 0.812\,$\pm$\,0.033 & 0.805\,$\pm$\,0.091 & 0.885\,$\pm$\,0.067 \\
XGB & 0.749\,$\pm$\,0.080 & 0.718\,$\pm$\,0.094 & 0.822\,$\pm$\,0.100 & 0.614\,$\pm$\,0.186 & 0.808\,$\pm$\,0.064 & 0.795\,$\pm$\,0.067 & 0.875\,$\pm$\,0.039 \\
DT  & 0.754\,$\pm$\,0.058 & 0.715\,$\pm$\,0.080 & 0.845\,$\pm$\,0.105 & 0.586\,$\pm$\,0.211 & 0.815\,$\pm$\,0.044 & 0.731\,$\pm$\,0.112 & 0.797\,$\pm$\,0.086 \\
MLP & 0.754\,$\pm$\,0.020 & 0.709\,$\pm$\,0.032 & 0.861\,$\pm$\,0.120 & 0.557\,$\pm$\,0.178 & 0.817\,$\pm$\,0.035 & 0.783\,$\pm$\,0.060 & 0.864\,$\pm$\,0.052 \\
LGBM& 0.724\,$\pm$\,0.067 & 0.702\,$\pm$\,0.071 & 0.776\,$\pm$\,0.136 & 0.629\,$\pm$\,0.185 & 0.781\,$\pm$\,0.068 & 0.783\,$\pm$\,0.073 & 0.857\,$\pm$\,0.039 \\
KNN & 0.694\,$\pm$\,0.044 & 0.633\,$\pm$\,0.031 & 0.838\,$\pm$\,0.099 & 0.429\,$\pm$\,0.087 & 0.778\,$\pm$\,0.044 & 0.702\,$\pm$\,0.048 & 0.796\,$\pm$\,0.051 \\
\bottomrule
\end{tabular}
\end{table}

\subsection{Model tuning: configuration analysis}
\label{sec:config}

Having compared the families, we now analyse which of the five configurations
each one selected and what that choice reveals about the data. The question is
not which setting is best in the abstract but how much model complexity 199
peptides can support. We examine this from two angles: the inner-loop capacity
curves in Table~\ref{tab:innercv}, and the distribution of selections across
folds in Figure~\ref{fig:selfreq}.

Table~\ref{tab:innercv} reports, for all 60 configurations, the balanced
accuracy achieved on the inner validation folds: each value is averaged over the
four inner folds and then over the five outer folds. Because the inner folds are
drawn from within each outer training partition, these are selection scores
rather than generalization estimates, but read as capacity curves they are
informative.
Three patterns emerge. The two families that perform best overall are also the
\emph{flattest}: LDA varies only between \num{0.719} and \num{0.733} across its
entire shrinkage range and LR between \num{0.683} and \num{0.730} across four
orders of magnitude of regularization, so the linear signal they capture is
robust rather than contingent on a fortunate penalty setting. Others show strong
monotone preferences that locate the small-sample regime precisely: SGD rises
sharply to \num{0.732} at its strongest penalty, LGBM declines monotonically as
its minimum leaf population grows, ADA rises monotonically and has not saturated
at 600 rounds, and XGB peaks at depth 1 --- a model with no interactions at all
--- and is flat thereafter. Finally, one configuration fails outright: SVM-S3
(radial-basis kernel, $C = 0.1$) scores exactly \num{0.500} in every inner fold,
its decision function being effectively constant. We keep such failures in the study
deliberately. Each family is swept from heavy regularization to almost none, and
a configuration that collapses at one end is evidence that the sweep is wide
enough to contain the useful settings rather than sampling a narrow band in the
middle of the range.

The four summary columns quantify these patterns. The highest family means
belong to LDA (\num{0.725}), LR (\num{0.712}) and ETC (\num{0.705}), the same
families that lead in Table~\ref{tab:fullmodels}. The standard deviations
separate the families whose performance depends on the swept parameter from
those where it does not: RFC (\num{0.003}), LDA (\num{0.005}) and XGB
(\num{0.006}) are essentially flat across their whole sweep, whereas SVM
(\num{0.088}), LGBM (\num{0.051}) and SGD (\num{0.040}) are not. The min--max
span makes the same point in the original units --- \num{0.014} for LDA against
\num{0.221} for SVM --- and identifies where tuning is worth the effort. For the
two best families it is not: LDA would have performed within \num{0.014}
balanced accuracy of its best whichever shrinkage value had been chosen.

\begin{table}[!htb]
\centering
\caption{Inner-loop balanced accuracy for each of the 60 configurations,
averaged over the four inner folds and then over the five outer folds (full
feature set). Best configuration per
family in bold. The parameter swept from S1 to S5 is given in
Table~\ref{tab:setups}. The last four columns summarize each family across its
own five configurations: a small standard deviation and a narrow min--max span
indicate a family whose performance is insensitive to the swept parameter.}
\label{tab:innercv}
\small
\setlength{\tabcolsep}{5pt}
\begin{tabular}{lccccc|cccc}
\toprule
Family & S1 & S2 & S3 & S4 & S5 & Mean & SD & Min & Max \\
\midrule
LR   & 0.716 & \textbf{0.730} & 0.725 & 0.706 & 0.683 & 0.712 & 0.019 & 0.683 & 0.730 \\
LDA  & 0.722 & 0.728 & \textbf{0.733} & 0.725 & 0.719 & 0.725 & \textbf{0.005} & 0.719 & 0.733 \\
SGD  & 0.649 & 0.641 & 0.661 & 0.632 & \textbf{0.732} & 0.663 & 0.040 & 0.632 & 0.732 \\
KNN  & 0.617 & 0.592 & 0.591 & 0.621 & \textbf{0.627} & 0.610 & 0.017 & 0.591 & 0.627 \\
DT   & 0.652 & 0.672 & \textbf{0.682} & 0.665 & 0.668 & 0.668 & 0.011 & 0.652 & 0.682 \\
SVM  & \textbf{0.721} & 0.649 & 0.500 & 0.693 & 0.692 & 0.651 & 0.088 & 0.500 & 0.721 \\
MLP  & 0.683 & 0.665 & 0.692 & \textbf{0.704} & 0.699 & 0.689 & 0.015 & 0.665 & 0.704 \\
RFC  & 0.687 & 0.693 & 0.693 & \textbf{0.694} & 0.694 & 0.692 & \textbf{0.003} & 0.687 & 0.694 \\
ETC  & 0.698 & \textbf{0.715} & 0.700 & 0.705 & 0.707 & 0.705 & 0.007 & 0.698 & 0.715 \\
ADA  & 0.583 & 0.629 & 0.664 & 0.672 & \textbf{0.681} & 0.646 & 0.040 & 0.583 & 0.681 \\
XGB  & \textbf{0.695} & 0.682 & 0.686 & 0.682 & 0.682 & 0.685 & 0.006 & 0.682 & 0.695 \\
LGBM & \textbf{0.689} & 0.684 & 0.665 & 0.593 & 0.582 & 0.643 & 0.051 & 0.582 & 0.689 \\
\bottomrule
\end{tabular}
\end{table}

To complement the previous results, Figure~\ref{fig:selfreq} shows which
configurations were labelled as the best estimator during the selection process.
Because this choice is made independently within each outer fold, a family may
select different configurations in different folds, and the spread of those
choices measures how stable the selection is. SGD selected its most strongly regularized configuration in all five
folds, the only unanimous choice in the study; SVM chose the linear kernel in
four of five folds and XGB the depth-1 setting in three of five. LDA's selection
instead spread across four of its five configurations, and KNN, RFC, ETC and
LGBM each split roughly evenly between two settings.

\begin{figure}[!htb]
\centering
\begin{tikzpicture}[x=1.05cm, y=0.46cm]
\foreach \fam [count=\r from 0] in
  {LR,LDA,SGD,KNN,DT,SVM,MLP,RFC,ETC,ADA,XGB,LGBM} {
  \node[font=\small, anchor=east] at (0.35,-\r) {\texttt{\fam}};
}
\foreach \s in {1,2,3,4,5} {
  \node[font=\small, anchor=south] at (\s,0.55) {S\s};
}
\foreach \row [count=\r from 0] in {%
  {1,3,0,1,0},  {1,1,2,0,1},  {0,0,0,0,5},  {2,0,0,1,2},
  {0,1,1,2,1},  {4,0,0,0,1},  {1,0,0,1,3},  {2,2,1,0,0},
  {2,2,0,1,0},  {0,0,1,1,3},  {3,0,1,0,1},  {2,1,2,0,0}} {
  \foreach \v [count=\c from 1] in \row {
    \pgfmathsetmacro{\shade}{\v*18}
    \pgfmathtruncatemacro{\shadeint}{\shade}
    \node[draw=black!35, line width=0.3pt, fill=black!\shadeint,
          minimum width=0.95cm, minimum height=0.42cm, inner sep=0pt,
          font=\small,
          text=\ifnum\v>3 white\else black\fi] at (\c,-\r) {\ifnum\v=0\relax\else\v\fi};
  }
}
\end{tikzpicture}
\caption{Configuration selected by the inner loop, counted over the five outer folds;
60 selections in total and each row sums to five. Darker cells mark more
frequent selection; empty cells were never selected.}
\label{fig:selfreq}
\end{figure}

This instability measures the data rather than the protocol. When five
configurations lie within \num{0.014} balanced accuracy of one another, as they
do for LDA, which one wins a given fold is close to arbitrary --- and that LDA
nonetheless achieves the best outer-fold performance indicates its result does
not depend on the shrinkage choice. The observation carrying most weight is the
convergence of SVM and XGB, from different model classes and through different
mechanisms, on configurations without non-linear interactions. Taken with the
flatness of the linear capacity curves, this is consistent evidence that 199
peptides support little model complexity --- which explains why the simpler
families lead in Section~\ref{sec:perf}.

\subsection{Model interpretation}
\label{sec:ablation}

Having compared the families and their configurations, we now ask \emph{which
features} the fitted models rely on. This subsection proceeds in three steps,
each motivated by the result of the previous one. We first apply SHAP to the
selected model of every family and find that the families divide into two
regimes that disagree completely about which feature matters most. We then test
that disagreement by removing the disputed feature and re-running the whole
protocol. Finally, because the two analyses do not agree, we introduce a null
test that determines which of them is informative.

\subsubsection{SHAP attribution for the selected models}
\label{sec:shapresults}

Following the protocol of Section~\ref{sec:shapmethod}, we applied SHAP to each
family at the configuration its inner loop had chosen most often across the five
folds, so that the model being explained is the one the protocol itself would
have selected.

Table~\ref{tab:shapshare} reports every feature's share of the total mean
absolute SHAP value, for all twelve families. The result is a clean split into
two regimes that disagree completely about which feature matters most. In the
five families that fit one coefficient per category level --- the three linear
models, the MLP and the SVM --- \texttt{GeneSymbol} dominates, taking between
\SI{68.7}{\percent} and \SI{81.7}{\percent} of the total attribution and
dwarfing every other feature, while HLA receives only
\SIrange{8.4}{14.3}{\percent}. The six tree ensembles show the exact opposite
profile: HLA takes \SIrange{36.9}{59.7}{\percent} and gene identity is assigned
essentially nothing, between \SI{0.0}{\percent} and \SI{3.7}{\percent}. KNN sits
between the two, spreading its attribution more evenly across all seven features
than any other family. Peptide length is the only other feature to attract a
substantial share, and it does so only in the tree ensembles, where it reaches
\SI{34.6}{\percent} for XGB.

\begin{table}[!htb]
\centering
\caption{Share of the total mean absolute SHAP value taken by each feature, per
family, at the configuration selected most often by the inner loop (full feature
set; rows sum to \SI{100}{\percent}). Families are grouped by how they represent
a categorical input. Largest share per row in bold.}
\label{tab:shapshare}
\small
\setlength{\tabcolsep}{4.5pt}
\begin{tabular}{l l ccccccc}
\toprule
Family & Setup & Gene & HLA & Tumor & Length & IsAnch. & mutPos. & RelPos. \\
\midrule
\multicolumn{9}{l}{\emph{one coefficient per category level}} \\
LDA  & LDA-S3  & \textbf{81.7} & 8.4  & 5.3  & 2.7  & 1.6  & 0.2  & 0.1 \\
MLP  & MLP-S5  & \textbf{71.8} & 13.0 & 6.5  & 4.2  & 3.6  & 0.2  & 0.7 \\
SVM  & SVM-S1  & \textbf{70.7} & 12.7 & 10.3 & 3.5  & 1.5  & 0.5  & 0.9 \\
SGD  & SGD-S5  & \textbf{69.1} & 13.5 & 8.0  & 4.9  & 3.4  & 0.7  & 0.3 \\
LR   & LR-S2   & \textbf{68.7} & 14.3 & 7.2  & 5.7  & 3.6  & 0.1  & 0.4 \\
\midrule
\multicolumn{9}{l}{\emph{tree-based}} \\
DT   & DT-S4   & 0.0 & \textbf{59.7} & 14.6 & 3.3  & 2.4  & 2.7  & 17.4 \\
RFC  & RFC-S2  & 2.1 & \textbf{48.2} & 11.8 & 21.8 & 7.0  & 5.1  & 4.0 \\
ETC  & ETC-S2  & 1.5 & \textbf{48.0} & 16.8 & 25.2 & 6.4  & 1.2  & 0.9 \\
ADA  & ADA-S5  & 3.7 & \textbf{47.9} & 11.6 & 30.4 & 6.4  & 0.0  & 0.0 \\
XGB  & XGB-S1  & 0.0 & \textbf{39.6} & 10.0 & 34.6 & 11.6 & 3.8  & 0.4 \\
LGBM & LGBM-S3 & 0.0 & \textbf{36.9} & 12.2 & 23.1 & 14.8 & 10.2 & 2.7 \\
\midrule
\multicolumn{9}{l}{\emph{instance-based}} \\
KNN  & KNN-S5  & 20.4 & \textbf{23.2} & 9.0 & 15.4 & 8.9 & 11.3 & 11.8 \\
\bottomrule
\end{tabular}
\end{table}

\begin{figure}[!htb]
\centering
\begin{minipage}{0.49\textwidth}
  \centering
  \includegraphics[width=\textwidth]{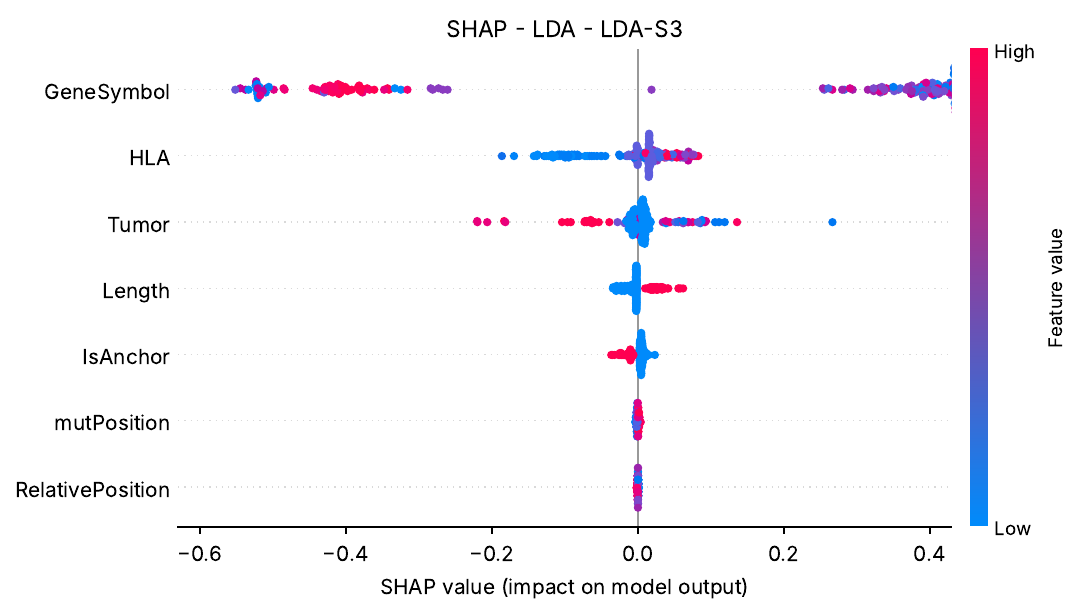}\\[2pt]
  {\small (a) LDA (setup LDA-S3)}
\end{minipage}\hfill
\begin{minipage}{0.49\textwidth}
  \centering
  \includegraphics[width=\textwidth]{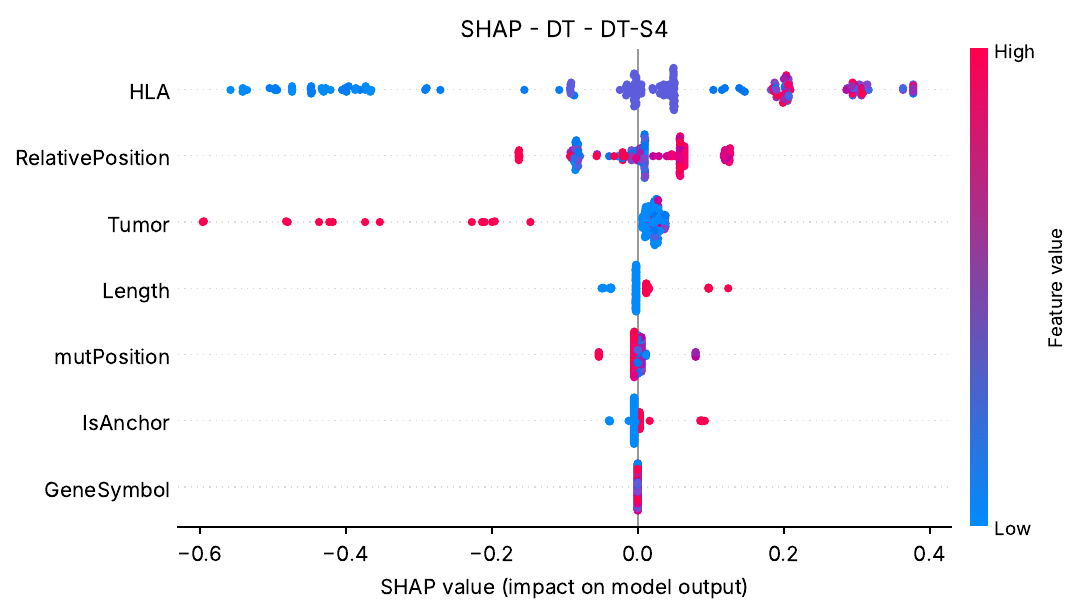}\\[2pt]
  {\small (b) DT (setup DT-S4)}
\end{minipage}
\caption{SHAP beeswarm summary plots for the two families whose attribution profiles
differ most: (a) LDA and (b) the decision tree. Rows are features ordered by
mean absolute SHAP value, most important at the top; each point is one of the
199 peptides, positioned by its SHAP value and coloured by the value of that
feature.}
\label{fig:beeswarm}
\end{figure}

The same contrast is visible at the level of individual peptides.
Figure~\ref{fig:beeswarm} shows the beeswarm summary plots for the two families
at the extremes of the split, LDA and the decision tree. Beyond the reversed
ordering, the two distributions have different shapes: LDA's gene attributions
are bimodal, splitting into two clusters of large magnitude with little mass in
between, whereas the decision tree spreads its HLA attributions continuously
across the whole range and assigns gene identity nothing at all.

Both regimes cannot be right about the same data. Either gene identity carries
most of the predictive signal and the tree ensembles are failing to exploit it,
or it does not and the linear and network models are being misled. The next
subsection settles this by removing the feature and measuring what happens.

\subsubsection{Feature ablation}
\label{sec:ablationresults}

We repeated the complete cross-validation protocol of
Section~\ref{sec:framework} under four feature sets: all features; all except
\texttt{GeneSymbol}; HLA alone; and all except \texttt{GeneSymbol} and HLA.
Table~\ref{tab:ablation} reports balanced accuracy for every family under each
of the four sets.

\begin{table}[!htb]
\centering
\caption{Mean outer-fold balanced accuracy of each family under the four feature sets.
``$-$Gene'' removes \texttt{GeneSymbol}; ``$-$Gene $-$HLA'' retains only the
tumor and structural features; ``HLA only'' retains the allele alone. Best value
per row in bold.}
\label{tab:ablation}
\small
\setlength{\tabcolsep}{8pt}
\begin{tabular}{lcccc}
\toprule
Family & All & $-$Gene & $-$Gene $-$HLA & HLA only \\
\midrule
LDA  & \textbf{0.760} & 0.713 & 0.649 & 0.681 \\
SGD  & 0.735 & \textbf{0.737} & 0.654 & 0.650 \\
ETC  & 0.733 & \textbf{0.736} & 0.684 & 0.701 \\
SVM  & \textbf{0.733} & 0.713 & 0.664 & 0.681 \\
RFC  & \textbf{0.730} & 0.722 & 0.666 & 0.701 \\
ADA  & 0.729 & \textbf{0.737} & 0.542 & 0.701 \\
LR   & 0.728 & \textbf{0.735} & 0.652 & 0.684 \\
XGB  & \textbf{0.718} & \textbf{0.718} & 0.659 & 0.701 \\
DT   & \textbf{0.715} & 0.701 & 0.619 & 0.697 \\
MLP  & 0.709 & \textbf{0.724} & 0.521 & 0.693 \\
LGBM & 0.702 & \textbf{0.713} & 0.650 & 0.696 \\
KNN  & 0.633 & 0.622 & 0.586 & \textbf{0.651} \\
\midrule
\textbf{Mean} & \textbf{0.719} & \textbf{0.714} (-0.7\,\%) & \textbf{0.629} (-12.5\,\%) & \textbf{0.686} (-4.6\,\%) \\
\bottomrule
\end{tabular}
\end{table}

Notice first that removing \texttt{GeneSymbol} changes nothing: the column mean
is essentially unaltered, from \num{0.719} to \num{0.714}, and removing the
feature is in fact the better condition for six of the twelve families. Removing
HLA as well costs every family instead. The third column keeps only the tumor
and structural features and the mean falls to \num{0.629}, the worst of the four
conditions; every family is worse off than in the second column, and two break
down altogether, MLP dropping to \num{0.521} and ADA to \num{0.542}, barely
above the \num{0.5} a model predicting a single class for every peptide would
obtain. Giving the models nothing but the allele recovers most of that loss but
not all of it: the fourth column reaches \num{0.686}, well above the
\num{0.629} obtained without HLA, yet still below the \num{0.714} obtained when
the tumor and structural features are present too. Taken together, the four
columns say that HLA is the feature that matters, that it is not sufficient on
its own, and that gene identity is dispensable --- which contradicts the SHAP
ranking of the previous subsection.

\subsubsection{A null test for the attribution}
\label{sec:null}

SHAP and the ablation disagree because they answer different questions. SHAP
asks how much an already fitted model's output moves when a feature is changed;
the ablation asks whether the model still performs as well when the feature is
removed and the model retrained. A feature can therefore earn a high SHAP value
and still be removable. What makes this particular disagreement informative is
which feature it affects. \texttt{GeneSymbol} records 173 genes among 199
peptides, so the model sees each gene roughly once and the weight it assigns to
any one gene rests on a single peptide; changing that gene moves the prediction
a great deal, which is exactly what SHAP measures. HLA, with 31 alleles and
about six peptides each, is in a different position. This suggests that the gene
attribution is an artifact of cardinality rather than evidence of signal, and
the present subsection tests that suggestion directly.

The explanation makes a testable prediction. If
\texttt{GeneSymbol} scores highly because it has many rarely seen categories,
and not because gene identity predicts immunogenicity, then a column with the
same 173 categories but no connection to the outcome should score just as
highly. We can build exactly that column: shuffle the genes at random among the
peptides. Every gene keeps its frequency, but which peptide carries which gene
is now arbitrary, so the column cannot possibly carry information. We refit each
model on the shuffled data and repeat the SHAP analysis, five times per family
(step 6 of the protocol in Section~\ref{sec:shapmethod}).

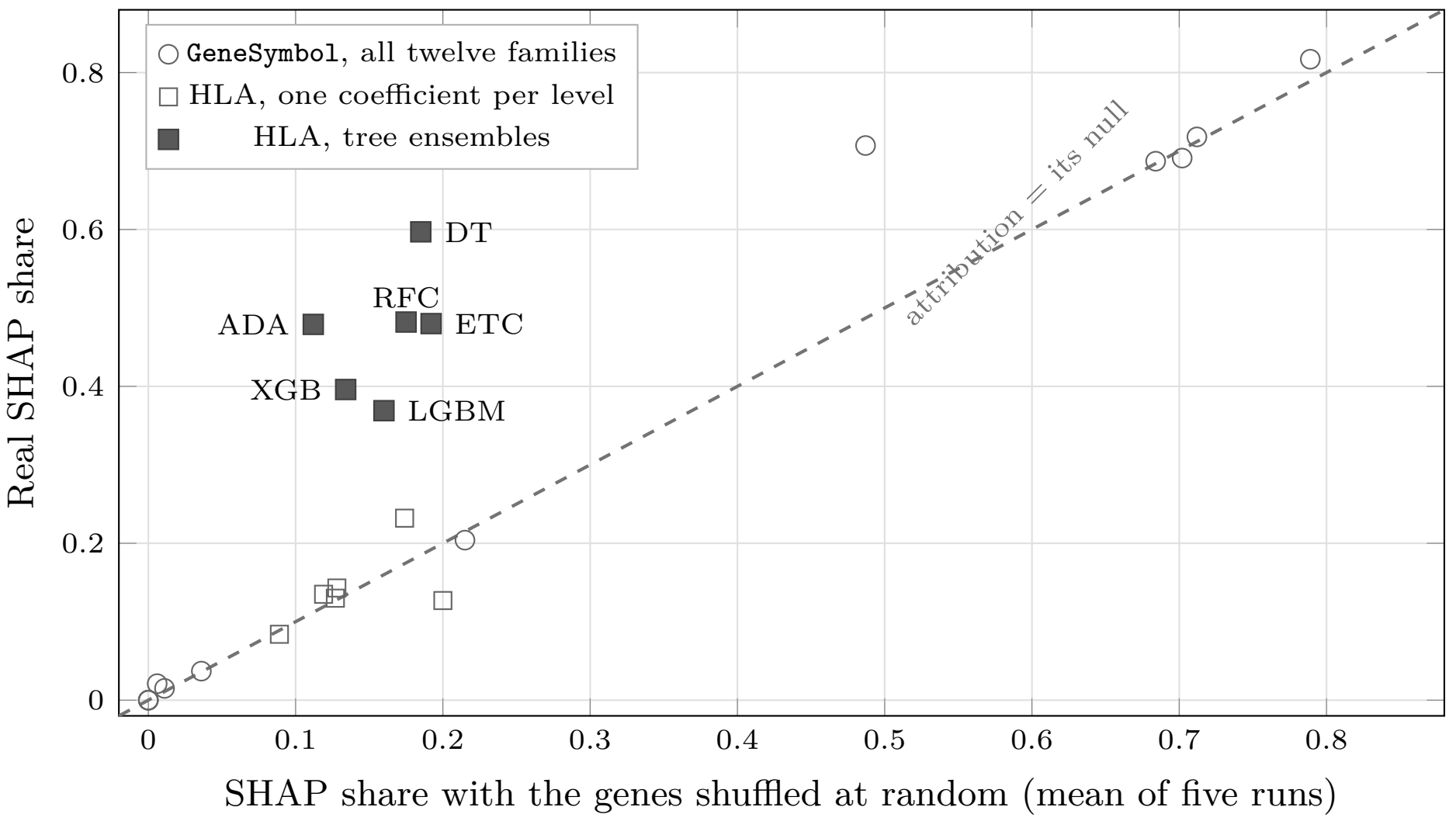
\begin{figure}[!htb]
\centering
\begin{tikzpicture}
\begin{axis}[
  width=0.78\textwidth, height=7.6cm,
  xlabel={SHAP share with the genes shuffled at random (mean of five runs)},
  ylabel={Real SHAP share},
  xlabel style={font=\small}, ylabel style={font=\small},
  xticklabel style={font=\scriptsize}, yticklabel style={font=\scriptsize},
  xmin=-0.02, xmax=0.88, ymin=-0.02, ymax=0.88,
  grid=major, grid style={black!12},
  legend style={at={(0.02,0.98)}, anchor=north west, font=\scriptsize,
                draw=black!30, fill=white},
]
\addplot[dashed, black!55, thick, domain=-0.02:0.88, samples=2, forget plot] {x};
\node[font=\scriptsize, text=black!55, rotate=45, anchor=south]
  at (axis cs:0.60,0.60) {attribution $=$ its null};

\addplot[only marks, mark=o, mark size=2.3pt, draw=black!60]
  coordinates {(0.684,0.687) (0.789,0.817) (0.702,0.691) (0.712,0.718)
               (0.487,0.707) (0.215,0.204) (0.000,0.000) (0.006,0.021)
               (0.011,0.015) (0.036,0.037) (0.000,0.000) (0.000,0.000)};
\addlegendentry{\texttt{GeneSymbol}, all twelve families}

\addplot[only marks, mark=square, mark size=2.1pt, draw=black!60]
  coordinates {(0.128,0.143) (0.089,0.084) (0.119,0.135) (0.127,0.130)
               (0.200,0.127) (0.174,0.232)};
\addlegendentry{HLA, one coefficient per level}

\addplot[only marks, mark=square*, mark size=2.4pt, draw=black!75, fill=black!65]
  coordinates {(0.185,0.597) (0.175,0.482) (0.192,0.480)
               (0.112,0.479) (0.134,0.396) (0.160,0.369)};
\addlegendentry{HLA, tree ensembles}
\node[font=\scriptsize, anchor=west]  at (axis cs:0.185,0.597) {\;DT};
\node[font=\scriptsize, anchor=south] at (axis cs:0.175,0.482) {RFC};
\node[font=\scriptsize, anchor=west]  at (axis cs:0.192,0.480) {\;ETC};
\node[font=\scriptsize, anchor=east]  at (axis cs:0.112,0.479) {ADA\;};
\node[font=\scriptsize, anchor=east]  at (axis cs:0.134,0.396) {XGB\;};
\node[font=\scriptsize, anchor=west]  at (axis cs:0.160,0.369) {\;LGBM};
\end{axis}
\end{tikzpicture}
\caption{Each feature's real SHAP share against the share it earns under the
shuffled-label control. All three categorical columns are shuffled together in
each control run; the figure shows the two whose attribution is in question,
\texttt{GeneSymbol} and \texttt{HLA}. Points on the dashed diagonal earn no more
than a meaningless column would; only distance above the diagonal can be
explained by genuine information.}
\label{fig:null}
\end{figure}

This gives each feature a reference value: the SHAP share it collects when it is
known to be meaningless. Figure~\ref{fig:null} plots each feature's real share
against that reference. A feature sitting on the diagonal earns no more than a
meaningless column would, and only the distance above the diagonal can be
attributed to genuine information.

The gene result does not survive the test. Of the five models that give every
gene its own weight, four show no gain at all: the shuffled column earns the
same SHAP share as the real one, the ratio of real to shuffled being \num{1.00}
for LR, \num{1.04} for LDA, \num{0.98} for SGD and \num{1.01} for MLP. In
Figure~\ref{fig:null} these points sit on the diagonal despite being the
highest-scoring points in the plot, so the \SIrange{68.7}{81.7}{\percent} share
that \texttt{GeneSymbol} obtained in Table~\ref{tab:shapshare} is reproduced in
full by a column known to carry no information whatsoever. It measures how many
genes there are, not how much they tell us. SVM, the fifth model in that group,
is the one exception (ratio \num{1.45}); its shuffled runs vary implausibly
little and we build no conclusion on it. KNN, which stores training examples
rather than fitting a weight per gene, also sits on the diagonal (ratio
\num{0.95}). The point is made most starkly by LDA, the best-performing model in
the whole study: not one of its seven features beats its shuffled counterpart,
HLA included.

The HLA result does survive it. In every tree-based model the real share exceeds
the shuffled one, by a factor of \num{2.31} (LGBM) to \num{4.30} (ADA), and
those six points sit well above the diagonal; averaged over the group the real
share is \num{3.01} times the shuffled share, against \num{1.03} for the models
that weight each gene separately. This is the only SHAP result in the study that
passes the test, it holds across six models built on different principles, and
it agrees with the ablation of Section~\ref{sec:ablationresults}, where removing
HLA cost every family.

The three steps therefore give one consistent account:
HLA restriction is the dominant feature, it is not sufficient on its own, and
the impression that gene identity dominated the linear and network models was an
artifact of having 173 genes for 199 peptides rather than a property of the
biology.

\section{Conclusions and Future Work}
\label{sec:conclusion}

This paper reported a systematic evaluation of 60 machine-learning
configurations for neoantigen immunogenicity prediction on experimentally
validated ITSNdb data, followed by a feature-ablation and interpretability
analysis. Our conclusions are:

\begin{enumerate}
  \item \textbf{A descriptive benchmark.} General-purpose classifiers reached
  balanced accuracies of \numrange{0.63}{0.76} and ROC-AUCs of
  \numrange{0.70}{0.82} under this protocol, with linear discriminant analysis
  highest on balanced accuracy ($0.760 \pm 0.100$) and a stochastic gradient
  linear model highest on ROC-AUC ($0.817 \pm 0.100$). The gradient-boosted ensembles attained lower mean
  balanced accuracy than the leading linear models while XGBoost retained
  competitive ranking performance; sample size is one possible contributor among
  several that this design does not separate.
  \item \textbf{Feature-set dependence.} Adding the HLA allele produced a
  positive estimated increment in balanced accuracy in 12 of 12 families, and adding the tumor and structural
  features on top of HLA improved 11 of 12, so HLA alone
  is not sufficient. Gene identity produced no detectable increment in any
  family.
  \item \textbf{Caution in interpretation.} A high SHAP share for gene
  identity (\SIrange{68.7}{81.7}{\percent} of the total in five families) did not by
  itself establish an incremental generalization benefit: a
  randomly scrambled copy of the same column, carrying no association with the
  outcome, earned the same share. SHAP values for one-hot categorical features
  in cohorts of this size should be reported against a shuffled-label control,
  and we describe one in full.
\end{enumerate}

Two directions for future work address the sample-size limitation underlying
several of these conclusions. \textbf{Scaling to a larger corpus}: the TumorAgDB platform
aggregates 1106 experimentally validated neoantigens spanning 15 cancer types or
cell lines \citep{shao2025}. Re-running this protocol there would roughly
quintuple the sample size, broaden allele and tumor coverage, and test whether
the gradient-boosted ensembles improve with scale. TumorAgDB assigns its
non-immunogenic label partly on an MHC binding affinity criterion
(IC\textsubscript{50} above \SI{500}{nM}) rather than purely on experimental
non-response, so the two resources are complementary rather than
interchangeable: ITSNdb offers stricter negatives, TumorAgDB greater scale.
\textbf{A systematic redundancy analysis}: the present study ablated features in
coarse groups, which separates uninformative from informative features but
cannot resolve redundancy among those retained. Quantifying feature--target and
feature--feature dependence under complementary criteria --- Pearson correlation,
mutual information for arbitrary and categorical dependence \citep{kraskov2004},
and the maximal information coefficient \citep{reshef2011} --- combined with
wrapper and embedded selection \citep{guyon2003}, would measure redundancy
directly. Wrapper selection scores a subset of features by the performance of a
model trained on it; embedded selection lets the model discard features as it
fits, as an L1 penalty does.

\sloppy

\end{document}